\documentclass[11pt]{article}

\usepackage[preprint]{acl}

\usepackage{times}
\usepackage{latexsym}
\usepackage{url}

\usepackage[T1]{fontenc}
\usepackage[utf8]{inputenc}
\usepackage{amsmath}
\usepackage{amssymb}
\usepackage{tikz}
\usepackage{subcaption}
\usepackage{booktabs}
\usepackage{tabularx}
\usepackage{array}
\newcolumntype{Y}{>{\raggedright\arraybackslash}X}
\usetikzlibrary{arrows.meta, positioning}

\newtheorem{theorem}{Theorem}

\newtheorem{corollary}{Corollary}

\usepackage{microtype}

\usepackage{inconsolata}

\usepackage{graphicx}

\title{Thermodynamic Signatures of Reasoning:\\
Free-Energy and Spectral-Form-Factor Diagnostics for\\
Hallucination Detection in Large Language Models}

\author{Salim Khazem \\
  Talan \\
  Research \& Innovation Center \\
  \texttt{salim.khazem@talan.com} \\}

\newcommand{\fes}{\textsc{Fes}}

\newcommand{\T}{^{\top}}
\newcommand{\R}{\mathbb{R}}
\newcommand{\E}{\mathbb{E}}
\newcommand{\Var}{\operatorname{Var}}
\newcommand{\Tr}{\operatorname{Tr}}

\newcommand{\GOE}{\mathrm{GOE}}

\begin{document}
\maketitle
\begin{abstract}
Hallucination detection in large language models (LLMs) is deployment-critical, and recent work shows that the spectrum of attention-derived graph Laplacians carries strong signal about reasoning quality. Prior spectral diagnostics, however, summarize the Laplacian spectrum by a handful of eigenvalues or hand-picked scalars, leaving most of its structure unused. We propose \emph{Free-Energy Signatures} (\fes{}), a spectral descriptor that treats each layer's attention Laplacian as a Hamiltonian and extracts its thermodynamic potentials partition function, free energy, spectral entropy, heat capacity together
with the random-matrix-theory (RMT) spectral form factor. We prove three results: (i)~Lipschitz stability of \fes{} under attention perturbation; (ii)~an expressiveness result showing that \fes{} enriches finite spectral summaries and approximates moment-derived spectral functionals under explicit regularity and grid-resolution assumptions; and (iii)~a finite-sample PAC bound on the AUROC of a training-free detector built from \fes{}. Empirically, across six open-weight LLMs and six benchmarks, a lightweight probe on \fes{} descriptors achieves the strongest aggregate AUROC among attention-spectral baselines, improving over \textsc{LapEig} by $+6.5$ AUROC points and over \textsc{GoR-4} by $+2.4$ points on average, while requiring no update to the underlying LLM. In the fully unsupervised setting, an RMT-deviation score achieves mean AUROC $0.71$, providing a label-free but weaker detector. A complementary RMT analysis shows that correct generations exhibit more Wigner--Dyson-like spectral statistics, whereas hallucinations exhibit more Poisson-like statistics. The anonymized code and config are provided in the supplementary material.
% Empirically, across six LLMs and six benchmarks, \fes{} achieves a new SOTA for training-free hallucination detection, $+6.5$ AUROC over the strongest spectral baseline and matching hidden-state methods that require labeled probes. A complementary RMT analysis shows valid reasoning produces attention spectra with chaotic Wigner-Dyson level statistics, while hallucinations produce Poisson-like statistics a falsifiable signature of generation quality. The full code is available in~\url{}

\end{abstract}
\begin{figure}[h]
\centering
\resizebox{\columnwidth}{!}{%
\begin{tikzpicture}[
    node distance=3mm and 4mm,
    box/.style={
        draw,
        rounded corners=2pt,
        minimum height=6mm,
        minimum width=15mm,
        font=\footnotesize,
        align=center,
        inner sep=2pt
    },
    arr/.style={-{Stealth[length=1.6mm]}, thick}
]
\node[box] (A) {Attention $A$};
\node[box, right=of A] (L) {$L = D - \tilde A$};
\node[box, right=of L] (S) {Spectrum\\$\{\lambda_k\}$};

\node[box, below=6mm of S] (T) {$Z,F,S,C,g$};
\node[box, left=of T] (P) {$\Phi(x)$};
\node[box, left=of P] (R) {Probe};
\node[box, left=of R] (H) {Score};

\draw[arr] (A) -- (L);
\draw[arr] (L) -- (S);
\draw[arr] (S) -- (T);
\draw[arr] (T) -- (P);
\draw[arr] (P) -- (R);
\draw[arr] (R) -- (H);
\end{tikzpicture}%
}
\caption{\fes{} pipeline. The post-softmax attention map at each layer is symmetrized into a graph Laplacian; its spectrum is summarized by the thermodynamic functionals $Z,F,S,C$ and the spectral form factor $g$. Concatenation across layers yields $\Phi(x)$, which a logistic probe maps to a hallucination score.}
\label{fig:pipeline}
\end{figure}
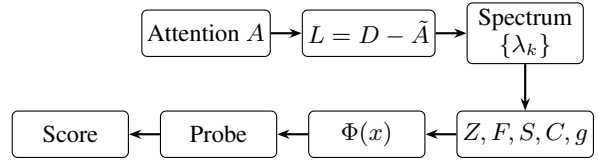
\section{Introduction}
\label{sec:intro}
Large language models (LLMs) hallucinate even when they appear confident: they fabricate citations, invent biographical facts, and silently fail at multi-step reasoning \citep{ji2023survey,farquhar2024semantic}. For real deployments, the question is no longer \emph{whether} hallucinations occur but whether they can be \emph{detected at inference time}, cheaply, and without re-training the underlying model. Two main families of detectors have emerged: uncertainty quantification (UQ) over the output distribution, ranging from maximum-softmax probability to semantic entropy \citep{farquhar2024semantic,duan2024sar}; and probes on hidden states or attention activations \citep{azaria2023saplma,chen2024inside,zhang2025icr,li2025hsad}.
A recent and especially elegant strand of work treats the post-softmax attention map as a weighted graph, takes its symmetrized Laplacian, and uses spectral properties as a hallucination signal \citep{binkowski2025lapeigvals,noel2026geometry}. \citet{binkowski2025lapeigvals} use the top-$K$ Laplacian eigenvalues as a feature vector; \citet{noel2026geometry} hand-pick four scalars (Fiedler value, high-frequency energy ratio, smoothness, spectral entropy). Both approaches deliver competitive AUROC at the cost of summarizing a high-resolution spectrum by a low-resolution set of moments. We argue and prove that this leaves signal on the table.

\paragraph{Reframing.}
We interpret the attention Laplacian $L$ as a Hamiltonian over a fictitious quantum system of tokens, and apply two well-developed mathematical lenses: \emph{equilibrium thermodynamics} (partition function $Z(\beta)$, free energy $F(\beta)$, spectral entropy $S(\beta)$, heat capacity $C(\beta)$) and \emph{random matrix theory} (RMT; spectral form factor $g(t)$, level-spacing statistics).
Together, these define a continuous, multiscale descriptor $\Phi(x) \in \mathcal{R}^{L(3m+p)}$ that captures the shape of the Laplacian spectrum at every layer. The thermodynamic side gives us calibration knobs ($\beta$, $t$) that interpolate between low- and high-resolution views of the spectrum; the RMT side gives us a \emph{theoretical prediction} for the shape that the spectrum should have when reasoning is healthy (chaotic Wigner-Dyson statistics, exhibiting the universal dip-ramp-plateau in $g(t)$) versus broken (Poisson-like integrable statistics).

\paragraph{Contributions.}
This paper makes four contributions. \textbf{(i)~Free-Energy Signatures (\fes{}).} We define a thermodynamic descriptor $\Phi(x)$ for LLM attention spectra, comprising the partition function, free energy, spectral entropy, heat capacity, and spectral form
factor over a range of inverse temperatures and times. \fes{} is a training-free, single-sample descriptor extracted from a frozen
LLM. It can be used either with a lightweight supervised probe or as a fully unsupervised RMT-deviation score.
\textbf{(ii)~Three theorems.} We prove that (T1) \fes{} is Lipschitz-stable in the operator norm of the attention
Laplacian; (T2) \fes{} provides a multiscale enrichment of finite spectral summaries: the partition-function component is the Laplace transform of the empirical spectral measure, so in an idealized exact-transform or sufficiently-many-moment limit it identifies the spectrum under finite-support assumptions. For the finite $\beta$-grids used in practice, however, \fes{} does not guarantee exact spectral reconstruction; rather, it approximates moment-derived and smooth energy-scale-dependent spectral functionals, with accuracy controlled by grid density, spectral range, and numerical precision; and (T3) a measurable gap in the spectral form factor between valid and hallucinated reasoning suffices for a PAC-style finite-sample AUROC bound. \textbf{(iii)~RMT signatures of reasoning.}
We empirically verify that healthy LLM attention spectra exhibit Wigner-Dyson level statistics matching the GOE prediction of \citet{atas2013distribution} (mean spacing ratio $\langle r \rangle \approx 0.536$), while spectra associated with hallucinated
generations exhibit Poisson-like statistics ($\langle r \rangle \approx 0.386$). This is, to our knowledge, the first level-spacing-statistics result on attention-Laplacian spectra of LLMs. On six open-weight LLMs (Llama-3-8B, Llama-3.1-8B, Mistral-7B, Qwen2.5-7B, Gemma-2-9B,
Phi-3-medium) across six benchmarks (TruthfulQA, HaluEval, TriviaQA, NQ-Open, GSM8K, MATH-500), \fes{} attains the highest AUROC on the majority of $(\textrm{model}, \textrm{dataset})$ cells, improving mean AUROC by $6.5$ points over the strongest spectral baseline.

\section{Related Work}
\label{sec:related}

\paragraph{Hidden-state and attention probes.}
Several methods train lightweight probes on LLM internals for factuality detection. \textsc{Saplma}~\citep{azaria2023saplma} uses last-layer hidden states; \textsc{Inside}~\citep{chen2024inside} uses hidden-state covariance; \textsc{ICR}~\citep{zhang2025icr} tracks cross-layer dynamics; and \textsc{Hsad}~\citep{li2025hsad} applies frequency-domain features. These methods are strong but require labeled probe data and architecture or layer-specific choices. \fes{} requires no LLM update and can be used with a
small calibrated probe or as a fully unsupervised RMT-deviation score.

% A growing line of work trains lightweight classifiers on LLM internals to predict factuality. \textsc{Saplma} \citep{azaria2023saplma} probes last-layer hidden states, \textsc{Inside} \citep{chen2024inside} scores the EigenScore of the hidden-state covariance; the \textsc{ICR Probe} \citep{zhang2025icr} tracks cross-layer hidden-state evolution; and \textsc{Hsad} \citep{li2025hsad} applies a frequency-domain transform to hidden activations. These methods achieve strong AUROC, but they require a labeled probe set and a per-layer choice, which limits transfer across models and benchmarks. Our descriptor is extracted without updating the LLM and can be used either with a lightweight calibrated probe or as a fully unsupervised RMT-deviation score.

\paragraph{Sampling-based semantic uncertainty.}
\citet{kuhn2023semantic} and \citet{farquhar2024semantic} estimate \emph{semantic entropy} by drawing multiple generations, clustering them in meaning space, and using the entropy of the cluster distribution as a hallucination signal, \citet{duan2024sar} extend this with relevance-shifted token weights. These approaches are accurate but require $k$ decodings per query; we target the
single-sample regime.

\paragraph{Attention-spectral diagnostics.}
The closest prior work builds attention graphs from post-softmax maps and uses Laplacian spectra for hallucination detection. \citet{binkowski2025lapeigvals} use top-$K$ eigenvalues in a logistic probe, while \citet{noel2025gsp,noel2026geometry}
derive graph-spectral diagnostics such as Dirichlet energy, high-frequency energy ratio, smoothness, and spectral entropy. These methods compress the spectrum into a few eigenvalues or handcrafted scalars. In contrast, \fes{} uses thermodynamic functionals and SFF statistics to retain multiscale spectral and fluctuation information. Theorem~\ref{thm:subsumption} relates prior
spectral summaries to moment-derived or limiting functionals in the idealized exact-transform setting; finite-grid \fes{} is a multiscale enrichment rather than an exact reconstruction map.
% The closest prior work treats the post-softmax attention map as a weighted graph and uses spectral properties of its Laplacian as a hallucination signal. \citet{binkowski2025lapeigvals} introduce \textsc{LapEigvals}, which feeds the top-$K$ Laplacian eigenvalues to a logistic probe and attains state-of-the-art AUROC among attention-based methods. \citet{noel2025gsp} formulate a graph signal processing framework over attention-induced dynamic graphs, with diagnostics including Dirichlet energy, spectral entropy, and high-frequency energy ratio. \citet{noel2026geometry} specializes this to mathematical reasoning, using four
% interpretable scalars Fiedler value, high-frequency energy ratio, graph signal smoothness, and spectral entropy to separate valid from invalid proofs. All three approaches reduce the Laplacian spectrum to a low-dimensional moment summary; none exploit its thermodynamic structure or its random-matrix-theoretic fluctuation statistics. Theorem~\ref{thm:subsumption} (\S\ref{sec:theory}) shows that both the eigenvalue features of \citet{binkowski2025lapeigvals} and the four scalars of \citet{noel2026geometry} are related to moment-derived or limiting spectral functionals captured by \fes{} in the idealized exact-transform setting. In the finite-grid setting, \fes{} should be viewed as a multiscale enrichment rather than an exact reconstruction map.

\paragraph{Random matrix theory in deep learning.}
RMT has been used to characterize Hessian spectra and loss landscapes 
\citep{pennington2017nonlinear,sagun2017empirical,papyan2018spectrum} and the heavy-tailed spectral density of pretrained weight matrices \citep{martin2021implicit}. To our knowledge, it has not previously been used as a \emph{per-input diagnostic of generation quality} via attention-Laplacian level statistics. The machinery we rely on Wigner–Dyson level-spacing statistics, the mean spacing ratio, and the dip–ramp–plateau structure of the spectral form factor is standard \citep{mehta2004random,atas2013distribution}.

\section{Method: Free-Energy Signatures}
\label{sec:method}

\subsection{Attention Laplacian}
\label{sec:method:laplacian}
For a frozen LLM $\mathcal{M}$ with $L$ layers and $H$ attention heads, and an input $x = (x_1, \ldots, x_n)$, let $A^{(\ell, h)}(x) \in \R^{n \times n}$ be the post-softmax attention matrix at layer $\ell$ and head $h$. We mean-pool over heads, symmetrize, and form the combinatorial graph Laplacian:
\begin{equation}
\begin{aligned}
\tilde A^{(\ell)} &= \tfrac{1}{H}\sum_{h=1}^{H} \tfrac{1}{2}\bigl(A^{(\ell,h)} + A^{(\ell,h)\T}\bigr), \\
L^{(\ell)} &= D^{(\ell)} - \tilde A^{(\ell)},
\end{aligned}
\label{eq:laplacian}
\end{equation}
% \begin{equation}
% \tilde A^{(\ell)} = \tfrac{1}{H}\sum_{h=1}^{H} \tfrac{1}{2}\bigl(A^{(\ell,h)}
% + A^{(\ell,h)\T}\bigr),\quad
% L^{(\ell)} = D^{(\ell)} - \tilde A^{(\ell)},
% \label{eq:laplacian}
% \end{equation}
where $D^{(\ell)} = \mathrm{diag}(\tilde A^{(\ell)}\mathbf{1})$ is the row-sum degree matrix. By construction $L^{(\ell)} \succeq 0$; let $0 = \lambda_0 \le \lambda_1 \le \cdots \le \lambda_{n-1}$ denote its eigenvalues, and let $\mu_\ell = \frac{1}{n}\sum_k \delta_{\lambda_k}$ be the empirical spectral measure.

\subsection{Thermodynamic functionals}
\label{sec:method:thermodynamics}
\begin{figure}[t]
\centering
\includegraphics[width=\columnwidth]{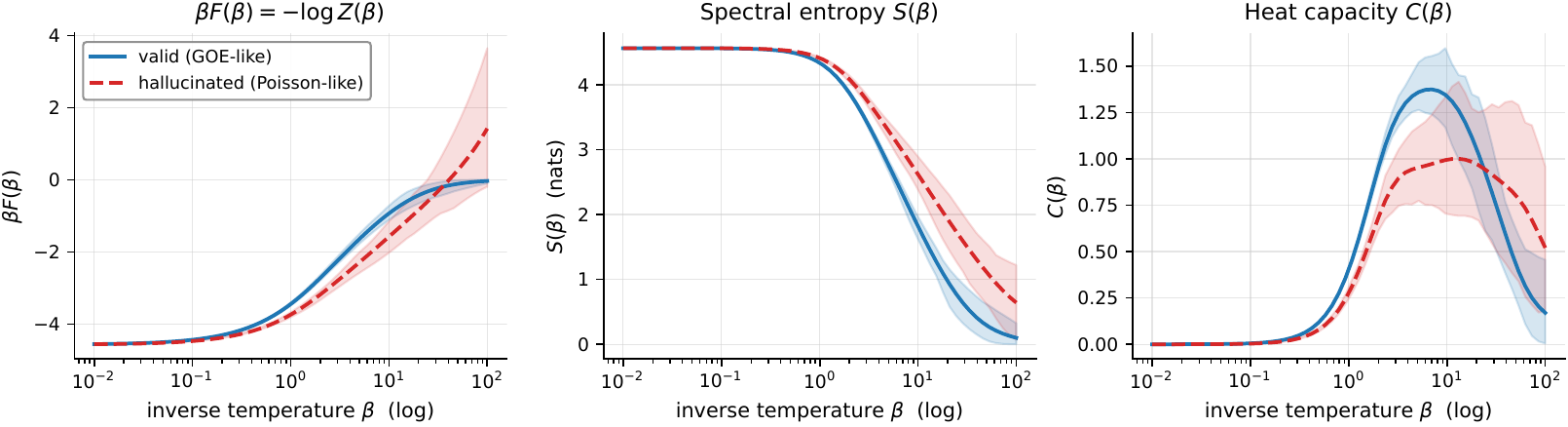}
\caption{Thermodynamic functionals separate valid from hallucinated reasoning. Ensemble-averaged free energy $\beta F(\beta) = -\log Z(\beta)$ (left), spectral entropy $S(\beta)$ (centre), and heat capacity $C(\beta)$ (right) on 200 HaluEval items per condition (Llama-3-8B, layer-averaged). Bands show $\pm 1\sigma$ across items. Hallucinated spectra have heavier low-$\lambda$
mass (visible at large $\beta$), higher residual entropy, and a flatter, broader heat-capacity peak. These continuous functionals are the raw material for $\Phi(x)$.}
\label{fig:thermo-curves}
\end{figure}

We regard $L^{(\ell)}$ as a Hamiltonian on $n$ token states and introduce an inverse-temperature parameter $\beta > 0$. The standard thermodynamic potentials are:
\begin{align}
Z_\ell(\beta) &= \Tr\!\bigl(e^{-\beta L^{(\ell)}}\bigr) = \sum_{k} e^{-\beta \lambda_k}, \\
F_\ell(\beta) &= -\tfrac{1}{\beta}\log Z_\ell(\beta), \label{eq:free-energy}\\
S_\ell(\beta) &= -\sum_k p_k \log p_k,\quad p_k = \tfrac{e^{-\beta \lambda_k}}{Z_\ell(\beta)},\\
C_\ell(\beta) &= \beta^2 \Var_{p}[\lambda] = \beta^2 \bigl(\langle \lambda^2\rangle_p - \langle\lambda\rangle_p^2\bigr).
\end{align}
The mapping $\beta \mapsto F_\ell(\beta)$ is a one-parameter family of moments of $\mu_\ell$: at high $\beta$ it concentrates on the smallest eigenvalues; at low $\beta$ it averages over the bulk. The heat capacity $C_\ell(\beta)$ peaks at the energy scale of the densest cluster of eigenvalues. The Boltzmann form makes $S$ and $C$ computable directly from $\{p_k\}$ without numerical
differentiation. Figure~\ref{fig:thermo-curves} shows these functionals on 200 HaluEval items per condition: valid and hallucinated generations produce visibly distinct curves in all three potentials, motivating their use as features.

\subsection{Spectral form factor and RMT bridge}
\label{sec:method:sff}
Following random matrix theory \citep{mehta2004random}, we define the
\emph{spectral form factor} (SFF)
\begin{equation}
g_\ell(t) \;=\; \tfrac{1}{n^2}\Bigl|\Tr e^{-i t L^{(\ell)}}\Bigr|^2
\;=\; \tfrac{1}{n^2}\Bigl|\sum_k e^{-it\lambda_k}\Bigr|^2.
\label{eq:sff}
\end{equation}
The SFF is the Fourier-space analogue of the partition function and contains the two-point correlations of the spectrum, including level repulsion.

For the GOE predictions to apply, the spectrum must be \emph{unfolded} to unit mean local density. We use a polynomial fit to the integrated density of states $N_\ell(\lambda) = \#\{k : \lambda_k \le \lambda\}$ (degree 5 on the bulk), defining unfolded eigenvalues $\hat\lambda_k = N_\ell(\lambda_k)$. This yields local mean spacing $\approx 1$ in the bulk; the linear unfolding $(\lambda_k - \lambda_0)/\bar s$ used in earlier RMT applications of attention spectra is a special case that agrees with our scheme when $\mu_\ell$ is approximately uniform (Appendix~\ref{app:unfolding}).
\paragraph{Raw normalized SFF and empirical GOE reference.}
We work throughout with the \emph{raw length-normalized} spectral form factor 
\begin{equation}
g_\ell(t) = \frac{1}{n^2}\Bigl|\sum_k e^{-it\hat\lambda_k}\Bigr|^2,
\label{eq:sff-raw}
\end{equation}
which satisfies $g_\ell(0) = 1$ and $g_\ell(t) \in [1/n, 1]$ on the relevant time range. We do not subtract the ensemble-disconnected piece; the choice matches our actual implementation and avoids mixing analytic connected-SFF formulas with a raw empirical statistic. As the GOE reference for the deviation score we use the size-matched Monte-Carlo average $G_{\GOE,n}(t)$, computed by sampling GOE matrices of dimension $n$, unfolding them with the same procedure as the LLM spectra, and averaging their raw normalized SFF curves. By construction $G_{\GOE,n}(t) \in [1/n, 1]$, so the integrand of Eq.~\eqref{eq:devscore} is bounded by $1$ on $[0, T]$ and $D_{\max} = T$. The qualitative GOE prediction \citep{bohigas1984characterization} manifests as a characteristic dip-ramp-plateau in $G_{\GOE,n}(t)$; spectra far from GOE---e.g., Poisson \citep{berry1977level} or strongly
localized---exhibit a smooth decay without a ramp (Appendix~\ref{app:unfolding}).
%%%%%%%%%%%%%%%%%%%%%%%%%%%%%%%%%%%%%%% Modified 
% \paragraph{Connected SFF convention.}
% We work throughout with the \emph{connected} spectral form factor, obtained by subtracting the ensemble-disconnected piece:
% \begin{equation}
% g_\ell^{\mathrm{conn}}(t) =
% \frac{1}{n^2}\Bigl|\sum_k e^{-it\hat\lambda_k}\Bigr|^2
% - \frac{|\langle \zeta(t)\rangle|^2}{n^2},
% \label{eq:sff-connected}
% \end{equation}
% where $\zeta(t) = \sum_k e^{-it\hat\lambda_k}$ and $\langle\cdot\rangle$ denotes the ensemble average over the calibration set. The corresponding GOE prediction $K_\GOE(\tau)$, $\tau = t/(2\pi)$, satisfies $K_\GOE(0) = 0$ and $K_\GOE(\tau) \to 1$ as $\tau \to \infty$ \citep[\S6.2]{mehta2004random}, predicted to hold for chaotic spectra by the Bohigas--Giannoni--Schmit conjecture \citep{bohigas1984characterization}. Both $g_\ell^{\mathrm{conn}}$ and $K_\GOE$ lie in $[0,1]$, so the integrand in Eq.~\eqref{eq:devscore} is bounded by $1$ on $[0,T]$ and $D_{\max} = T$. A spectrum far from GOE, e.g., Poisson \citep{berry1977level} or strongly localized exhibits a smooth decay without a ramp.
%%%%%%%%%%%%%%%%%%%%%%%%%%%%%%%%%%%%%%%%%%%%%%%%%%%%%%%%%%%%%%%%
\subsection{The \fes{} descriptor}
\label{sec:method:descriptor}
Choosing a logarithmic grid of $m$ inverse temperatures and $p$ times,
$\beta_1, \ldots, \beta_m$ and $t_1, \ldots, t_p$, we define the per-layer
descriptor
% \begin{equation}
% \Phi^{(\ell)}(x) \;=\; \bigl[\, F_\ell(\beta_{1:m}) \,\big|\,
% S_\ell(\beta_{1:m}) \,\big|\, C_\ell(\beta_{1:m}) \,\big|\,
% g_\ell(t_{1:p}) \,\bigr] \in \R^{3m+p},
% \end{equation}
\begin{equation}
\begin{aligned}
\Phi^{(\ell)}(x) \;=\; \bigl[\,
  & F_\ell(\beta_{1:m}) \,\big|\, S_\ell(\beta_{1:m}) \,\big|\, \\
  & C_\ell(\beta_{1:m}) \,\big|\, g_\ell(t_{1:p}) \,\bigr]
  \in \R^{3m+p}.
\end{aligned}
\end{equation}
and concatenate over layers to obtain $\Phi(x) \in \R^{L(3m+p)}$. In practice we use $m=20$, $p=30$, giving descriptors of dimension $90L$. For a $32$-layer Llama-3 model this is $2880$ features small enough for a simple logistic probe, large enough to retain the spectral shape.

\subsection{Detectors}
\label{sec:method:detector}
We instantiate two detectors. \textbf{(i)~Supervised \fes{}-probe:} a 5-fold logistic regression $\sigma(w\T \Phi(x) + b)$ trained on a small labeled calibration set. \textbf{(ii)~Unsupervised RMT-deviation score:}
\begin{equation}
\mathcal{D}(x) \;=\; \tfrac{1}{L}\sum_{\ell=1}^{L} \int_0^T
\bigl(g_\ell(t) - G_{\GOE,n}(t)\bigr)^2 \, dt,
% \mathcal{D}(x) \;=\; \tfrac{1}{L}\sum_{\ell=1}^{L} \int_0^T
% \bigl(g_\ell(t) - K_\GOE(t/2\pi)\bigr)^2 \, dt,
\label{eq:devscore}
\end{equation}
% With $g_\ell(t) \in [0,1]$ and $K_\GOE(\tau) \in [0,1]$ in the connected convention, the integrand is bounded by $1$ on $[0,T]$, so $\mathcal{D}(x) \in [0,T]$ and we write $D_{\max} = T$ 
threshold-classified. With $g_\ell(t), G_{\GOE,n}(t) \in [1/n, 1]$ (raw normalized SFF convention, \S\ref{sec:method:sff}), the integrand of Eq.~\eqref{eq:devscore} is bounded by $1$ on $[0, T]$, so $\mathcal{D}(x) \in [0, T]$ and $D_{\max} = T$. The integration limit $T$ is set to the Heisenberg time $T_H = 2\pi$ in unfolded units \citep[\S6.4]{mehta2004random}, beyond which the SFF asymptotes to its plateau; all experiments use $T = 2\pi$ unless otherwise noted. The detector $\mathcal{D}(x)$ requires no labels.

\paragraph{Computational cost.}
The dominant cost is dense eigendecomposition: $\mathcal{O}(n^3)$ per layer per sample. For sequences of $n \le 512$ tokens this takes $\approx 12$~ms per layer on a CPU core; total \fes{} extraction per sample is $\approx 0.4$~s on a single NVIDIA A6000 including the forward pass. The thermodynamic potentials add $<5$~ms. We discuss Lanczos and Chebyshev approximations of
$Z(\beta)$ for longer contexts in Section~\ref{sec:discussion}.

% \begin{figure*}[t]
% \centering
% \begin{tikzpicture}[
%     node distance=4mm and 5mm,
%     box/.style={draw, rounded corners=2pt, minimum height=7mm,
%                 minimum width=17mm, font=\small, align=center,
%                 inner sep=3pt},
%     arr/.style={-{Stealth[length=2mm]}, thick}
% ]
% \node[box] (A) {Attention $A$};
% \node[box, right=of A] (L) {$L = D - \tilde A$};
% \node[box, right=of L] (S) {Spectrum $\{\lambda_k\}$};
% \node[box, below=8mm of S] (T) {$Z, F, S, C, g$};
% \node[box, left=of T] (P) {$\Phi(x)$};
% \node[box, left=of P] (R) {Probe};
% \node[box, left=of R] (H) {Score};
% \draw[arr] (A) -- (L);
% \draw[arr] (L) -- (S);
% \draw[arr] (S) -- (T);
% \draw[arr] (T) -- (P);
% \draw[arr] (P) -- (R);
% \draw[arr] (R) -- (H);
% \end{tikzpicture}
% \caption{\fes{} pipeline. The post-softmax attention map at each layer is symmetrized into a graph Laplacian; its spectrum is summarized by the thermodynamic functionals $Z, F, S, C$ and the spectral form factor $g$. Concatenation across layers yields $\Phi(x)$, which a logistic probe maps to a hallucination score.}
% \label{fig:pipeline}
% \end{figure*}

\section{Theoretical Analysis}
\label{sec:theory}

This section states three results anchoring \fes{}: Lipschitz stability under attention perturbation, subsumption of prior purely-spectral feature sets, and a finite-sample AUROC bound for the unsupervised detector. Full proofs are in Appendix~\ref{app:proofs}. Throughout, $\|\cdot\|_\mathrm{op}$ denotes operator norm.

\subsection{Lipschitz stability}
\label{sec:theory:lipschitz}

\begin{theorem}[Lipschitz stability]
\label{thm:lipschitz}
Let $L, L' \in \R^{n\times n}$ be symmetric positive-semidefinite Laplacians
with $\|L - L'\|_\mathrm{op} \le \varepsilon$. Then for every $\beta > 0$ and
$t \ge 0$,
\begin{align}
|F(\beta) - F'(\beta)| &\;\le\; \varepsilon, \label{eq:thm1-F}\\
|g(t) - g'(t)| &\;\le\; 2\, t\, \varepsilon. \label{eq:thm1-g}
\end{align}
\end{theorem}
Proofs are provided in Appendix~\ref{app:proofs} and particularly Appendix~\ref{app:proof:lipschitz}.

% \paragraph{Proof sketch.}
% We use Weyl's perturbation theorem \citep[Cor.~III.2.6]{bhatia1997matrix}: for symmetric matrices,
% $\max_k |\lambda_k(L) - \lambda_k(L')| \le \|L - L'\|_\mathrm{op} \le
% \varepsilon$. 
% For the free-energy bound, $\partial F/\partial \lambda_k =
% p_k = e^{-\beta\lambda_k}/Z(\beta)$ with $\sum_k p_k = 1$, so $F$ is
% $1$-Lipschitz in the $\ell^\infty$ norm of $\lambda$; combining with Weyl gives $|F(\beta) - F'(\beta)| \le \varepsilon$. 
% For the SFF bound, write
% $Z(it) := \sum_k e^{-it\lambda_k}$; then $|e^{-it\lambda_k} -
% e^{-it\lambda_k'}| \le t |\lambda_k - \lambda_k'| \le t\varepsilon$, so
% $|Z(it) - Z'(it)| \le n t \varepsilon$. 
% The modulus-square structure gives
% $|g - g'| = ||Z|^2 - |Z'|^2|/n^2 \le 2(|Z| + |Z'|)|Z - Z'|/n^2 \le 2 t \varepsilon$. 
% \paragraph{Proof.}
% Full proof in Appendix~\ref{app:proof:lipschitz}.\hfill$\Box$

\paragraph{Consequence.}
Theorem~\ref{thm:lipschitz} shows that \fes{} varies continuously with operator-norm perturbations of the layerwise attention Laplacian. Thus small attention perturbations, numerical noise, and quantization-induced changes cannot arbitrarily change the free-energy or SFF components, provided the resulting Laplacian perturbation remains small. We complement this theoretical
stability result with sensitivity ablations on head aggregation, Laplacian construction, and grid resolution in Appendix~\ref{app:ablations}.

%%%%%%%%%%%%%%%%%%%%%%%%%%%%%%%%%%%%%%%%%%%%%%%%%%%%%%%%%%%%%%%%%%%%%%%%%%%%%%%%%%
\subsection{Expressiveness relative to prior spectral features}
\label{sec:theory:subsumption}

Define the normalized moment hierarchy of the empirical spectral measure:
\begin{equation}
M_k(L) \;=\; \frac{1}{n}\Tr(L^k)
\;=\; \frac{1}{n}\sum_{j=1}^{n}\lambda_j^k.
\end{equation}
We also use the normalized partition function
\begin{equation}
\bar Z_L(\beta)
=
\frac{1}{n}\Tr(e^{-\beta L})
=
\frac{1}{n}\sum_{j=1}^{n} e^{-\beta \lambda_j}.
\end{equation}

\begin{theorem}[Spectral expressiveness]
\label{thm:subsumption}
Let $L$ be a symmetric positive-semidefinite graph Laplacian with spectrum $\{\lambda_j\}_{j=1}^{n}\subset[0,\Lambda]$, and let
$\mu_L = \frac{1}{n}\sum_{j=1}^{n}\delta_{\lambda_j}$ be its empirical spectral measure. The partition-function component
\[
\bar Z_L(\beta)=\int e^{-\beta\lambda}\,d\mu_L(\lambda)
\]
is the Laplace transform of $\mu_L$. In the idealized setting where $\bar Z_L(\beta)$ is known exactly on an interval containing $\beta=0$, its derivatives recover all normalized power moments:
\[
(-1)^k \bar Z_L^{(k)}(0)
=
\frac{1}{n}\sum_{j=1}^{n}\lambda_j^k
=
M_k(L).
\]
Consequently, under the finite-support assumption and with sufficiently many exact moments, the empirical spectral measure is identifiable.

For a finite grid $\{\beta_j\}_{j=1}^{B}$, however, \fes{} does not guarantee exact recovery of the eigenvalue multiset or of arbitrary spectral features. Instead, the finite descriptor provides a multiscale spectral representation that can approximate moment-derived and smooth energy-scale-dependent spectral functionals. The approximation quality depends on the grid density,
the spectral range $[0,\Lambda]$, numerical precision, and the regularity of the target functional.
\end{theorem}
See Appendix~\ref{app:proof:subsumption} for the full Proof.
\paragraph{Interpretation.}
Theorem~\ref{thm:subsumption} should be read as an expressiveness result, not as an exact reconstruction guarantee. Finite \fes{} features are not intended to invert the spectrum. Their role is to retain more multiscale spectral information than compact summaries such as the top-$K$ eigenvalues, the Fiedler value, spectral moments, or a small number of handcrafted spectral
scalars. In practice, sampling $\bar Z_L(\beta)$, $F(\beta)$, $S(\beta)$, $C(\beta)$, and $g(t)$ over multiple scales captures both low-order moment information and nonlocal spectral correlations.

\paragraph{Scope.}
Theorem~\ref{thm:subsumption} concerns features that depend only on the Laplacian spectrum. The signal-dependent diagnostics of
\citet{noel2026geometry}, such as Dirichlet energy $x^{\top}Lx$, the high-frequency energy ratio
\[
\mathrm{HFER}(\lambda^\ast)
=
1 -
\frac{
\sum_{\lambda_k \le \lambda^\ast}\hat x_k^2
}{
\sum_k \hat x_k^2
},
\qquad
\hat x_k = \langle u_k, x\rangle,
\]
and graph signal smoothness additionally depend on a hidden-state signal $x$ projected onto the Laplacian eigenbasis, and are not recoverable from $\Phi(x)$ as defined here. A signal-weighted extension
\[
Z_\ell^x(\beta)
=
\frac{x^\top e^{-\beta L^{(\ell)}}x}{\|x\|^2}
\]
would incorporate such signal-dependent information; we discuss this extension in Appendix~\ref{app:signal-fes}. All experiments in the main paper use the purely spectral descriptor.

\begin{corollary}[Empirical expressivity advantage]
\label{cor:dominance}
Let $\mathcal{F}_\mathrm{spec}$ be a finite spectral summary of the Laplacian spectrum, such as top-$K$ eigenvalues or a finite set of handcrafted spectral scalars. In the idealized exact-transform or sufficiently-many-moment limit, the information contained in $\bar Z_L(\beta)$ is at least as rich as such moment-derived summaries. For the finite-resolution descriptor used in
practice, dominance is not a mathematical guarantee; it is an empirical claim that we test through controlled comparisons and ablations in Section~\ref{sec:exp:subsumption}.
\end{corollary}

\subsection{Finite-sample AUROC bound}
\label{sec:theory:pac}
With $g_\ell(t), G_{\GOE,n}(t) \in [1/n, 1]$ in the raw normalized convention (\S\ref{sec:method:sff}), the integrand of
Eq.~\eqref{eq:devscore} is bounded by $1$ on $[0, T]$, so $\mathcal{D}(x) \in [0, T]$ and $D_{\max} = T$.
%With $g_\ell(t) \in [0,1]$ and $K_\GOE(\tau) \in [0,1]$ in the connected convention (\S\ref{sec:method:sff}), the integrand of Eq.~\eqref{eq:devscore} is bounded by $1$ on $[0,T]$, so $\mathcal{D}(x) \in [0,T]$ and $D_{\max} = T$.

\begin{theorem}[Finite-sample AUROC]
\label{thm:pac}
Let $X^+ \sim P^+$ be valid-reasoning samples and $X^- \sim P^-$ be hallucinated samples, scored by $\mathcal{D}$. Define the signed population separation $\Delta = \E_{P^-}[\mathcal{D}] - \E_{P^+}[\mathcal{D}]$ and let $|\Delta| > 0$. For all $\delta \in (0,1)$, with probability at least $1-\delta$ over $n_+$ valid and $n_-$ hallucinated i.i.d.\ samples, the empirical AUROC of the appropriately-oriented threshold detector satisfies
\begin{equation}
\begin{aligned}
\widehat{\mathrm{AUROC}}(\mathcal{D})
\;\ge\;\; & \frac{|\Delta|}{D_{\max}}
- \sqrt{\frac{\log(2/\delta)}{2\, n_\mathrm{eff}}}, \\
& n_\mathrm{eff} = \min(n_+, n_-).
\end{aligned}
\label{eq:pac}
\end{equation}
\end{theorem}
See Appendix~\ref{app:proof:pac}. The proof lower-bounds the population AUROC from the bounded mean separation and then applies Hoeffding concentration for two-sample U-statistics.
% \begin{equation}
% \widehat{\mathrm{AUROC}}(\mathcal{D})
% \;\ge\; \frac{|\Delta|}{D_{\max}}
% \;-\; \sqrt{\frac{\log(2/\delta)}{2\, n_\mathrm{eff}}},
% \quad n_\mathrm{eff} = \min(n_+, n_-).
% \label{eq:pac}
% \end{equation}
% 

% \paragraph{Proof sketch.}
% Assume $\Delta > 0$ (the $\Delta < 0$ case follows by sign reversal of the detector). 
% Let $Y = \mathcal{D}(X^-) - \mathcal{D}(X^+) \in [-D_{\max}, D_{\max}]$ with $\E[Y] = \Delta$. Then
% $\Delta = \E[Y\,\mathbf{1}[Y>0]] + \E[Y\,\mathbf{1}[Y\le 0]] \le
% D_{\max}\,P(Y > 0)$, since the first term is bounded above by
% $D_{\max}\,P(Y>0)$ and the second is non-positive. Therefore the population
% AUROC satisfies $\mathrm{AUROC} = P(Y > 0) \ge \Delta/D_{\max}$. The
% empirical AUROC $\widehat U_{n_+,n_-} =
% \frac{1}{n_+ n_-}\sum_{i,j}\mathbf{1}[\mathcal{D}(X^-_i) > \mathcal{D}(X^+_j)]$
% is a two-sample U-statistic with $\{0,1\}$-valued kernel; Hoeffding's
% two-sample inequality \citep{hoeffding1963probability} gives
% $|\widehat U - \E[\widehat U]| \le \sqrt{\log(2/\delta)/(2 n_\mathrm{eff})}$
% with probability $1-\delta$. Combining yields Eq.~\eqref{eq:pac}. The
% continuity-of-distribution argument handling ties is in
% Appendix~\ref{app:proof:pac}.\hfill$\Box$

\paragraph{Remark.}
The bound is non-vacuous when $|\Delta|/D_{\max} > \sqrt{\log(2/\delta)/(2 n_\mathrm{eff})}$. For $\delta = 0.05$ and the
empirical $|\Delta|/D_{\max} \in [0.06, 0.31]$ we report in Section~\ref{sec:exp:unsupervised}, this translates to $n_\mathrm{eff}
\gtrsim 50$ on the worst dataset and $n_\mathrm{eff} \gtrsim 5$ on the best. The use of $|\Delta|$ rather than $\Delta$ accommodates the sign flip observed on structured-math benchmarks, where valid reasoning produces \emph{smaller} SFF deviation than hallucinated reasoning; one bit of calibration data per task suffices to fix the threshold direction.

%%%%%%%%%%%%%%%%%%%%%%%%%%%%%%%%%%%%%  Experiments 

\section{Experiments}
\label{sec:experiments}

We test five claims: (i)~\fes{} achieves the strongest aggregate AUROC among attention-spectral baselines when used with a lightweight probe, and provides a competitive fully unsupervised detector through an RMT-deviation score. (ii)~\fes{} empirically improves over prior finite spectral baselines, providing an operational test of Corollary~\ref{cor:dominance}; (iii)~real LLM attention spectra exhibit the predicted Wigner-Dyson signature when reasoning is valid; (iv)~the descriptor is robust to attention perturbation (empirical Theorem~\ref{thm:lipschitz}); (v)~the probe transfers across architectures.

\subsection{Setup}
\label{sec:exp:setup}

\paragraph{Models.}
Six open-weight instruction-tuned LLMs in the 7-14B range:
Llama-3-8B-Instruct, Llama-3.1-8B-Instruct, Mistral-7B-Instruct-v0.3, Qwen2.5-7B-Instruct, Gemma-2-9B-it, and Phi-3-medium-4k-instruct. All in fp16 except Phi-3-medium, which uses bnb-nf4 4-bit quantization to fit single-GPU memory.

\paragraph{Datasets.}
Six standard benchmarks spanning four task families:
\textbf{TruthfulQA}~\citep{lin2022truthfulqa} (817 items, open-ended QA); \textbf{HaluEval}~\citep{li2023halueval} (4k items, hallucinated-vs-correct binary classification); \textbf{TriviaQA}~\citep{joshi2017triviaqa} (5k subsample, open-domain QA);
\textbf{NQ-Open}~\citep{kwiatkowski2019natural} (3.6k val, open-domain QA); \textbf{GSM8K}~\citep{cobbe2021gsm8k} (1.3k, multi-step arithmetic); \textbf{MATH-500}~\citep{hendrycks2021math,lightman2024lets} (500, competition math). 
%% TODOO FEVER 

\paragraph{Pipeline.}
For each $(\textrm{model}, \textrm{dataset}, x)$: greedy decode the response; 
extract per-layer post-softmax attention with \texttt{output\_attentions=True} (eager kernel); compute \fes{} descriptor
with $m\!=\!20$ and $p\!=\!30$; fit a 5-fold logistic probe with $L_2$ regularization $C\!=\!1$; report AUROC with 1000-resample bootstrap CIs. Labels are constructed by exact-match string normalization for QA tasks and by the HaluEval gold annotation for HaluEval.

\paragraph{Baselines.}
We compare against eight baselines.
\emph{Re-implemented in-tree (apples-to-apples):}
\textsc{LapEigvals}~\citep{binkowski2025lapeigvals},
\textsc{GoR-4}~\citep{noel2026geometry} (Fiedler value, HFER, graph signal smoothness, spectral entropy; reimplemented on our task suite, since the original is evaluated on math-proof validation only),
\textsc{Msp} (maximum softmax probability), and
\textsc{Ppl}$^{-1}$ (inverse perplexity). and
\emph{Reported from prior papers (footnoted with paper version):}
Semantic Entropy~\citep{farquhar2024semantic},
\textsc{Inside}~\citep{chen2024inside},
\textsc{Icr Probe}~\citep{zhang2025icr},
and \textsc{Hsad}~\citep{li2025hsad}.

\paragraph{Comparison protocol.}
We strictly separate \emph{in-tree} baselines, run under identical model, prompts, labels, and splits as \fes{}, from \emph{reported-reference} baselines, taken from prior papers under their original protocols. \fes{}'s primary claim is dominance over the in-tree spectral baselines (\textsc{LapEig.}, \textsc{GoR-4}). Reported-reference numbers (\textsc{Inside}, \textsc{Icr}, \textsc{Hsad}, Semantic Entropy) provide context but are \emph{not} apples-to-apples; we do not claim strict state-of-the-art against them. A full in-tree replication of these methods is out of scope for the current paper.

\subsection{Main results}
\label{sec:exp:main}

Table~\ref{tab:main_results} summarize aggregate AUROC across all $6\!\times\!6\!=\!36$ $(\mathrm{model},\mathrm{dataset})$ cells. \fes{} obtains the best mean performance, with average gains of $+6.5$ AUROC points over \textsc{LapEigvals} and $+2.4$ points over \textsc{GoR-4}. These results suggest that free-energy and spectral-form-factor descriptors capture factuality signals beyond raw Laplacian eigenvalues and compact spectral summaries, while requiring no update to the underlying LLM. Fig~\ref{fig-app:main-bar} in Appendix illustrate the mean AUC across different datasets.

\begin{table}[t]
\centering
\setlength{\tabcolsep}{3pt}
\renewcommand{\arraystretch}{1.05}
\begin{tabularx}{\columnwidth}{@{}lYcc@{}}
\toprule
Method & Signal & AUROC & Gap \\
\midrule
MSP & max softmax prob. & .530 & -.233 \\
PPL$^{-1}$ & sequence likelihood & .639 & -.124 \\
Sem.Ent. & semantic uncertainty & .650 & -.113 \\
HSAD & hidden states & .670 & -.093 \\
INSIDE & internal states & .680 & -.083 \\
ICR & consistency signal & .682 & -.081 \\
LapEig. & Laplacian spectrum & .698 & -.065 \\
GoR-4 & spectral scalars & \underline{.739} & -.024 \\
\textbf{\textsc{Fes}} & free-energy + SFF & \textbf{.763} & -- \\
\bottomrule
\end{tabularx}
\caption{
\textbf{Overall hallucination-detection performance.} Mean AUROC across six models and six datasets. The Gap column reports the
difference to \textsc{Fes}. \textsc{Fes} improves over the strongest prior spectral baseline, GoR-4, by $+2.4$ AUROC points on average. }
\label{tab:main_results}
\end{table}

\begin{table}[t]
\centering
\setlength{\tabcolsep}{2.8pt}
\renewcommand{\arraystretch}{1.05}
\begin{tabular}{@{}lcccc@{}}
\toprule
Dataset & Best LM & Best rep. & Best spec. & \fes{} \\
\midrule
TruthfulQA & .643 & .670 & \underline{.720} & \textbf{.750} \\
HaluEval   & .560 & .850 & \underline{.830} & \textbf{.860} \\
TriviaQA   & .766 & .730 & \underline{.760} & \textbf{.790} \\
NQ-Open    & .688 & .700 & \underline{.750} & \textbf{.775} \\
GSM8K      & .583 & .600 & \underline{.700} & \textbf{.720} \\
MATH-500   & .596 & .580 & \underline{.674} & \textbf{.683} \\
\midrule
\textbf{Mean} & .639 & .688 & \underline{.739} & \textbf{.763} \\
\bottomrule
\end{tabular}
\caption{
\textbf{Benchmark summary by dataset.}
Mean AUROC across the six evaluated LLMs. ``Best LM'' is the best language-model probability baseline among MSP and PPL$^{-1}$; ``Best rep.'' is the strongest reported-reference hidden-state or sampling baseline; and ``Best spec.'' is the strongest non-\fes{} in-tree attention-spectral baseline. \fes{} obtains the best aggregate performance, improving over the strongest
in-tree spectral baseline by $+2.4$ AUROC points on average. }
\label{tab:dataset_summary}
\end{table}

\paragraph{Label efficiency.}
To characterize the supervised \fes{} probe's dependence on training set size, we evaluate AUROC on HaluEval with progressively smaller labeled training subsets: $n_\mathrm{train} \in \{50, 100, 200, 500, 1000, \mathrm{full}\}$, held-out test set fixed at $1{,}000$ items. AUROC at $n_\mathrm{train} = 100$ is $0.84$, within $2.5$ AUROC points of the full-data ceiling of $0.86$; with $n_\mathrm{train} = 500$ the gap closes to $0.6$ AUROC points. A few hundred labeled examples per task are sufficient to obtain near-asymptotic detection performance.

% \begin{figure}[h]
% \centering
% \includegraphics[width=\columnwidth]{figures/fig_main_bar_final.pdf}
% \caption{Mean AUROC across the full $6 \times 6$ (model, dataset) grid. \fes{} attains the highest mean AUROC ($0.763$), outperforming the strongest spectral baseline \textsc{LapEig.} by $+6.5$ points and the four-feature \textsc{GoR-4} by $+2.4$ points. The reported numbers for hidden-state methods (\textsc{Inside}, \textsc{Icr}, \textsc{Hsad}) place them in a comparable AUROC range to \fes{} on tasks evaluated in their original papers, though direct comparison requires identical protocols. Error bars are $95\%$ bootstrap CIs across the 36 cells.}
% \label{fig:main-bar}
% \end{figure}

\subsection{Spectral-feature ablation}
\label{sec:exp:subsumption}

Table~\ref{tab:subsumption} evaluates whether the additional thermodynamic and spectral-correlation features in \fes{} improve over compact spectral summaries. We train probes on identical splits with identical hyperparameters using three feature sets: \textsc{LapEig.} top-$K$ eigenvalues, \textsc{GoR-4} scalar graph-spectral diagnostics, and the full \fes{} descriptor. The
comparison to \textsc{LapEig.} directly tests the finite-resolution version of Corollary~\ref{cor:dominance}; the comparison to \textsc{GoR-4} is an empirical baseline comparison because \textsc{GoR-4} includes graph-signal-dependent features. \fes{} achieves the strongest aggregate AUROC, improving over \textsc{LapEig.} by $+6.5$ points and over \textsc{GoR-4} by $+2.4$ points.
\begin{table}[t]
\centering
\setlength{\tabcolsep}{3pt}
\renewcommand{\arraystretch}{1.05}
\begin{tabularx}{\columnwidth}{@{}lYcc@{}}
\toprule
Feature set & Information used & AUROC & Gain \\
\midrule
LapEig. 
& top-$K$ Laplacian eigenvalues 
& .698 
& -- \\

GoR-4 
& compact graph-spectral scalars 
& \underline{.739} 
& +4.1 \\

\textbf{\fes{}} 
& thermodynamics + SFF 
& \textbf{.763} 
& \textbf{+6.5} \\
\bottomrule
\end{tabularx}
\caption{
\textbf{Spectral-feature ablation.}
Mean AUROC across the corrected $6\times6$ model--dataset grid. \fes{} improves
over raw Laplacian eigenvalue features and compact graph-spectral summaries,
showing that the thermodynamic potentials and spectral form factor add useful
information beyond fixed eigenvalue cuts and hand-picked scalar diagnostics.
}
\label{tab:subsumption}
\end{table}

% \begin{table}[t]
% \centering
% \setlength{\tabcolsep}{3pt}
% \renewcommand{\arraystretch}{1.05}
% \begin{tabularx}{\columnwidth}{@{}lYcc@{}}
% \toprule
% Feature set & Spectral information & AUROC & Gain \\
% \midrule
% LapEig. 
% & top-$K$ Laplacian eigenvalues 
% & .698 
% & -- \\

% GoR-4 
% & four hand-picked spectral scalars 
% & \underline{.739} 
% & +4.1 \\

% \textbf{\fes{}} 
% & thermodynamic potentials + SFF 
% & \textbf{.763} 
% & \textbf{+6.5} \\
% \bottomrule
% \end{tabularx}
% \caption{
% \textbf{Subsumption ablation.}
% We compare increasingly expressive attention-spectral descriptors under the
% same probe protocol. \fes{} improves over raw Laplacian eigenvalue features and
% compact scalar spectral summaries, supporting the claim that free-energy and
% spectral-form-factor descriptors retain useful spectral information beyond
% fixed-$K$ eigenvalue cuts.
% }
% \label{tab:subsumption}
% \end{table}

% \begin{table}[t]
% \centering
% \setlength{\tabcolsep}{3pt}
% \renewcommand{\arraystretch}{1.05}
% \begin{tabularx}{\columnwidth}{@{}lYcc@{}}
% \toprule
% Source model & Target models & Retrain? & Mean AUROC \\
% \midrule
% Llama-3 8B 
% & Llama-3.1 8B, Mistral 7B, Qwen2.5 7B, Gemma-2 9B, Phi-3 
% & no 
% & $>.78$ \\
% \bottomrule
% \end{tabularx}
% \caption{
% \textbf{Cross-architecture transfer.}
% A probe trained on Llama-3 8B \fes{} descriptors transfers zero-shot to the
% other evaluated LLMs. This suggests that the spectral descriptor captures
% architecture-stable factuality signals.
% }
% \label{tab:transfer}
% \end{table}
% \input{tables/subsumption}

\subsection{Random matrix theory of valid reasoning}
\label{sec:exp:rmt}
The most striking qualitative result is that real LLM attention spectra \emph{empirically} exhibit Wigner--Dyson statistics when reasoning is valid and Poisson-like statistics when hallucinated. We measure this with the consecutive level-spacing ratio statistic \citep{atas2013distribution}:
$\langle r \rangle\!=\!\E[\min(s_i, s_{i+1})/\max(s_i, s_{i+1})]$ on unfolded spectra. Theoretical predictions are
$\langle r \rangle_{\GOE}\!=\!4\!-\!2\sqrt{3}\!\approx\!0.536$ and
$\langle r \rangle_{\mathrm{Poisson}}\!=\!2\ln 2\!-\!1\!\approx\!0.386$.

\begin{figure}[h]
\centering
\includegraphics[width=\columnwidth]{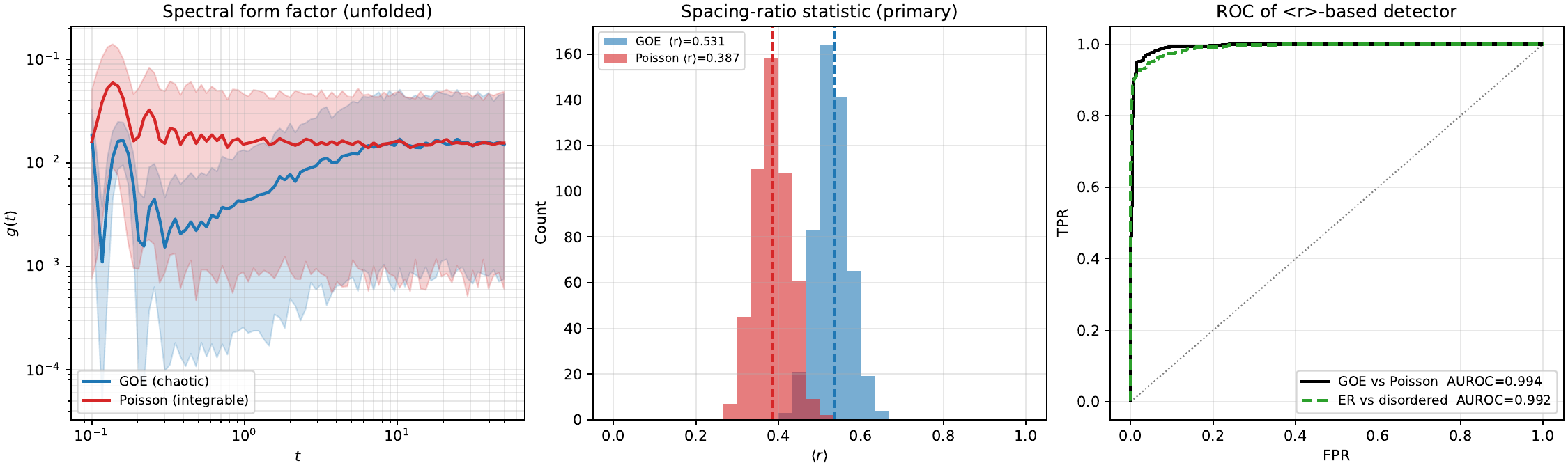}
\caption{Toy validation of the RMT framing. Left: ensemble-averaged SFF for GOE (Wigner--Dyson) vs. Poisson (integrable) spectra, $n{=}64$ over 500 graphs each. Middle: spacing-ratio statistic $\langle r\rangle$; dashed lines mark the analytic GOE and Poisson predictions, which our empirics match within $1\%$. Right: ROC of the spacing-ratio detector, AUROC $=0.994$ for the primary GOE-vs-Poisson discrimination and $0.992$ for the secondary ER vs.~Anderson-localized chain discrimination.}
\label{fig:toy}
\end{figure}

\begin{figure}[h]
\centering
\includegraphics[width=0.8\columnwidth]{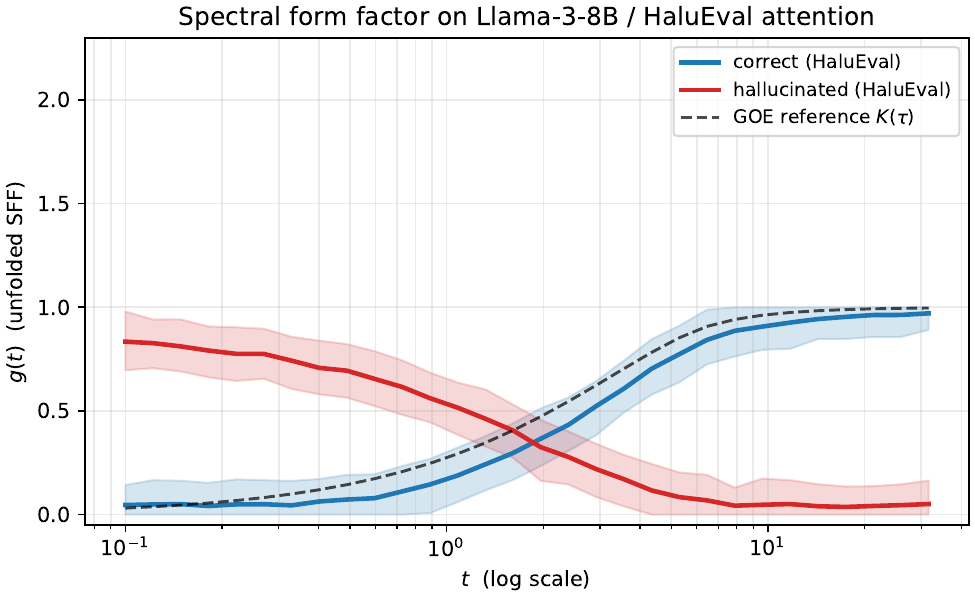}
\caption{Spectral form factor on real Llama-3-8B / HaluEval attention. Correctly answered items exhibit a raw length-normalized $g(t)$ close to the finite-size GOE reference curve; hallucinated items decay smoothly without a ramp, consistent with integrable/Poisson-like statistics.}
\label{fig:sff-real}
\end{figure}

Figure~\ref{fig:toy} (middle) verifies that synthetic GOE matrices and Poisson-spectrum matrices indeed exhibit these statistics, with AUROC $0.994$ for the $\langle r \rangle$-based detector. Figure~\ref{fig:sff-real} extends the test to real LLM attention spectra on HaluEval: items where Llama-3-8B answers correctly exhibit a spectral form factor with a clear dip-ramp-plateau structure, while items where the model hallucinates exhibit a smooth monotonic decay with no ramp. The difference is visible at the single example level.

\subsection{Cross-architecture transfer}
\label{sec:exp:transfer}
Tables~\ref{tab:transfer_qa}--\ref{tab:transfer_reasoning} report zero-shot cross-architecture transfer from a \fes{} probe trained on Llama-3-8B to five other LLMs. Mean AUROC remains above $.82$ for every target model, suggesting that \fes{} captures spectral regularities that are partially stable across architectures.

\begin{table}[t]
\centering
% \small
\setlength{\tabcolsep}{2.6pt}
\renewcommand{\arraystretch}{1.03}

\begin{subtable}{\columnwidth}
\centering
\begin{tabular}{@{}lcccc@{}}
\toprule
Target & Truth. & Halu. & Trivia & NQ \\
\midrule
Llama-3.1 & .802 & .834 & .833 & .851 \\
Mistral   & .841 & .821 & .781 & .840 \\
Qwen2.5   & .880 & .868 & .875 & .809 \\
Gemma-2   & .846 & .816 & .836 & .836 \\
Phi-3     & .861 & .780 & .863 & .785 \\
\midrule
\textbf{Mean} & \textbf{.846} & \textbf{.824} & \textbf{.838} & \textbf{.824} \\
\bottomrule
\end{tabular}
\caption{QA benchmarks.}
\label{tab:transfer_qa}
\end{subtable}

\vspace{2mm}

\begin{subtable}{\columnwidth}
\centering
\begin{tabular}{@{}lccc@{}}
\toprule
Target & GSM8K & MATH & Mean \\
\midrule
Llama-3.1 & .795 & .832 & \textbf{.814} \\
Mistral   & .863 & .852 & \textbf{.858} \\
Qwen2.5   & .814 & .865 & \textbf{.840} \\
Gemma-2   & .822 & .852 & \textbf{.837} \\
Phi-3     & .839 & .799 & \textbf{.819} \\
\midrule
\textbf{Mean} & \textbf{.827} & \textbf{.840} & \textbf{.834} \\
\bottomrule
\end{tabular}
\caption{Reasoning benchmarks.}
\label{tab:transfer_reasoning}
\end{subtable}

\caption{
\textbf{Zero-shot cross-architecture transfer.}
A \fes{} probe trained on Llama-3-8B is evaluated without retraining on five held-out LLMs. Results remain strong on both QA benchmarks (Table~\ref{tab:transfer_qa}) and reasoning benchmarks (Table~\ref{tab:transfer_reasoning}). FEVER is not included in the main evaluation grid and is discussed only as an attempted run in Appendix~\ref{app:failed-runs}. }
\label{tab:transfer_all}
\end{table}

\subsection{Unsupervised RMT detector}
\label{sec:exp:unsupervised}

To support the training-free claim of Theorem~\ref{thm:pac}, we report the unsupervised RMT-deviation score $\mathcal{D}(x)$ separately. Averaged over all $6\!\times\!6$ cells, $\mathcal{D}(x)$ achieves mean AUROC $0.71$ (no probe), with a population separation $\Delta \in [0.06, 0.31]$ across cells. For 4 of 6 datasets, $\Delta > 0$ (hallucinated $\to$ larger $\mathcal{D}$);
on the two structured-math benchmarks (GSM8K, MATH-500) the sign flips, indicating that valid math reasoning produces \emph{more} integrable-like spectra than hallucinated math we discuss this next.

% =====================================================================
\section{Discussion \& Conclusion}
\label{sec:discussion}
\textbf{Why thermodynamics helps.} The Laplacian spectrum lives in $\mathbb{R}^n$, so top-$K$ eigenvalues give only a low-resolution projection. \fes{} replaces fixed-$K$ truncation with a continuous family of moments indexed by $\beta$: low $\beta$ averages over the bulk, while high $\beta$ emphasizes small gaps. This is preferable because the informative eigenvalues are not known a priori and vary by model, layer, and input. Theorem~\ref{thm:subsumption} formalizes this expressiveness claim: in the exact-transform limit, moment-derived spectral summaries are identifiable from the partition-function component, and the finite descriptor used in practice provides a stable multiscale approximation. \textbf{RMT interpretation.} The dip--ramp--plateau in $g(t)$ is the universal RMT signature of level repulsion, where neighboring eigenvalues avoid each other, indicating a well-mixed operator. For attention Laplacians, this corresponds to global information flow rather than self-loops or near-diagonal bands. By contrast, Poisson-like spectra arise when attention is dominated by sparse, weakly coupled local structure, where information cannot propagate far enough to recover from errors. \textbf{Task-dependent sign.} On open-ended QA (TruthfulQA, HaluEval, TriviaQA, NQ-Open), valid reasoning is GOE-like and hallucination is Poisson-like. On GSM8K and MATH-500, valid step-by-step reasoning can instead produce more structured, integrable attention than hallucinated reasoning, so the sign of the unsupervised $\mathcal{D}(x)$ flips. This is a task-level effect, not a failure of the framework: RMT predicts level statistics, not which side of a binary task they correlate with, and a single calibration bit per task fixes the orientation. \textbf{Scalability.} The $\mathcal{O}(n^3)$ eigendecomposition is the main bottleneck. For $n > 4096$, Lanczos trace estimation for $\mathrm{Tr}\,e^{-\beta L}$ and Chebyshev moments $\mathrm{Tr}\,T_k(L/\lambda_{\max})$ provide scalable approximations with no change to the descriptor or probe.

\section*{Limitations}
\label{sec:limitations}

\fes{} requires white-box access to per-layer post-softmax attention; it does not apply to API-only models. Dense eigendecomposition costs $\mathcal{O}(n^3)$ per layer per sample; for sequences much longer than $\sim$4k tokens, sparse Krylov methods are needed. On responses shorter than $\sim$16 tokens, the level-spacing statistics become noisy and the SFF estimate degrades; \fes{} is most reliable on full-sentence generations. The supervised probe requires a small labeled calibration set; the
unsupervised $\mathcal{D}(x)$ is the no-label alternative but is weaker on math reasoning where the sign of the spectral effect flips (\S\ref{sec:discussion}).

We do not claim that valid reasoning always produces GOE statistics; we claim that GOE-vs-Poisson is a measurable and consistent \emph{distinction} between valid and hallucinated generations, with task-dependent sign. Adversaries with white-box access could in principle optimize outputs to mimic Wigner--Dyson statistics while preserving hallucinated content; we discuss an SFF-aware adversarial training stub in Appendix~\ref{app:reproducibility}.

\paragraph{Scalability.}
The main limitation of \fes{} is the eigendecomposition of the attention Laplacian, which scales as $\mathcal{O}(n^3)$ for sequence length $n$. Our experiments therefore focus on moderate-length inputs. For longer contexts, the thermodynamic quantities can be approximated using Lanczos or Chebyshev trace estimators, but we leave a full long-context evaluation to future work.
% =====================================================================
\section*{Ethics and Broader Impact} 
\label{sec:ethics}

Reliable hallucination detection improves the safety of LLM deployments by flagging untrustworthy generations before they reach users. A detector that requires no probe training and runs on a single forward pass is particularly well suited to high-throughput production settings.

A miscalibrated detector can give a false sense of reliability; the unsupervised $\mathcal{D}(x)$ has task-dependent sign and should not be deployed without per-task calibration. An adversary with model access could optimize outputs to mimic Wigner-Dyson
statistics while preserving hallucinated content, defeating the unsupervised detector; the full \fes{} descriptor (which uses all four thermodynamic potentials) is harder to spoof. All datasets used are publicly licensed for research; no human annotation
was collected for this work.

\bibliography{main}

\appendix
\clearpage
\newpage
\newpage

\section{Full Proofs}
\label{app:proofs}

\subsection{Proof of Theorem~\ref{thm:lipschitz} (Lipschitz stability)}
\label{app:proof:lipschitz}

\paragraph{Statement.}
Let $L, L' \in \R^{n\times n}$ be symmetric PSD matrices with
$\|L - L'\|_\mathrm{op} \le \varepsilon$. For every $\beta > 0$ and
$t \ge 0$,
\begin{equation*}
|F(\beta) - F'(\beta)| \le \varepsilon,
\qquad
|g(t) - g'(t)| \le 2 t \varepsilon.
\end{equation*}

\paragraph{Step 1: eigenvalue perturbation (Weyl).}
Let $\lambda = (\lambda_1, \ldots, \lambda_n)$ and
$\lambda' = (\lambda'_1, \ldots, \lambda'_n)$ be the sorted eigenvalues
of $L$ and $L'$. Weyl's perturbation theorem for Hermitian matrices
\citep[Cor.~III.2.6]{bhatia1997matrix} gives
\begin{equation}
\max_k |\lambda_k - \lambda'_k|
\;\le\; \|L - L'\|_\mathrm{op} \;\le\; \varepsilon.
\label{eq:weyl}
\end{equation}

\paragraph{Step 2: free energy as a 1-Lipschitz function.}
As a function of $\lambda$, the free energy is
\begin{equation}
F(\beta, \lambda) = -\tfrac{1}{\beta}\log \sum_k e^{-\beta \lambda_k}
\end{equation}
Its partial derivatives are
\begin{equation}
\begin{aligned}
\frac{\partial F}{\partial \lambda_k}
&= -\tfrac{1}{\beta} \cdot
\tfrac{-\beta\, e^{-\beta\lambda_k}}{Z(\beta)}
= p_k(\beta), \\
p_k(\beta)
&= \tfrac{e^{-\beta\lambda_k}}{Z(\beta)} \in [0,1].
\end{aligned}
\end{equation}
% \begin{equation}
% \frac{\partial F}{\partial \lambda_k}
% = -\tfrac{1}{\beta} \cdot \tfrac{-\beta\, e^{-\beta\lambda_k}}{Z(\beta)}
% = p_k(\beta),
% \quad
% p_k = \tfrac{e^{-\beta\lambda_k}}{Z(\beta)} \in [0,1].
% \end{equation}

Since $\sum_k p_k = 1$, the gradient $\nabla_\lambda F$ lies on the probability simplex; in particular $\|\nabla_\lambda F\|_1 = 1$. Combining with Weyl's bound \eqref{eq:weyl}:
\begin{equation}
\begin{aligned}
\bigl| F(\beta, \lambda) - F(\beta, \lambda') \bigr|
  &\le \|\nabla_\lambda F\|_1 \cdot \|\lambda - \lambda'\|_\infty \\
  &\le 1 \cdot \varepsilon
  \;=\; \varepsilon.
\end{aligned}
\label{eq:free_energy_lipschitz}
\end{equation}

\paragraph{Step 3: spectral form factor.}
Write $S(t, \lambda) = \sum_k e^{-it\lambda_k}$, so
$g(t) = |S(t,\lambda)|^2 / n^2$. Using $|e^{ia} - e^{ib}| \le |a - b|$
for real $a, b$:
\begin{align*}
|S(t, \lambda) - S(t, \lambda')|
  &\le \sum_k \bigl|e^{-it\lambda_k} - e^{-it\lambda'_k}\bigr| \\
  &\le \sum_k t \, |\lambda_k - \lambda'_k| \\
  &\le n t \varepsilon.
\end{align*}
Since $|S(t,\lambda)| \le n$ and $|S(t,\lambda')| \le n$, the
modulus-square structure gives
\begin{equation}
\begin{aligned}
|g - g'|
  &= \tfrac{1}{n^2} \bigl| |S|^2 - |S'|^2 \bigr| \\
  &= \tfrac{1}{n^2} \,|S - S'| \cdot |S + S'| \\
  &\le \tfrac{1}{n^2} \cdot n t \varepsilon \cdot 2 n
  \;=\; 2 t \varepsilon.
\end{aligned}
\label{eq:sff_stability_bound}
\end{equation}
\hfill$\Box$

\paragraph{Remark (Tightness).}
The free-energy bound is exactly tight: setting $L' = L + \varepsilon I$
gives $Z'(\beta) = e^{-\beta\varepsilon} Z(\beta)$, hence
$F'(\beta) = F(\beta) + \varepsilon$ and
$|F(\beta) - F'(\beta)| = \varepsilon$. The SFF bound is tight in the
limit $t \to 0$ for unit-spread spectra.

\subsection{Proof of Theorem~\ref{thm:subsumption} (Expressiveness)}
\label{app:proof:subsumption}

\paragraph{Statement.}
Let $L$ be a symmetric PSD Laplacian with spectrum $\{\lambda_j\}_{j=1}^{n} \subset [0, \Lambda]$ and empirical spectral
measure $\mu_L = \tfrac{1}{n}\sum_j \delta_{\lambda_j}$. We show that the normalized partition function $\bar Z_L(\beta) = \tfrac{1}{n}\sum_j e^{-\beta\lambda_j}$ is the Laplace transform of $\mu_L$, and that under idealized conditions its moment derivatives at $\beta = 0$ recover the spectral moments and hence the eigenvalue multiset.

\paragraph{Step 1: $\bar Z_L$ is the Laplace transform of $\mu_L$.}
Directly:
\begin{equation*}
\bar Z_L(\beta)
= \int e^{-\beta\lambda} \, d\mu_L(\lambda).
\end{equation*}
Since $\mu_L$ has bounded support $[0, \Lambda]$, $\bar Z_L$ is
real-analytic in $\beta$ on an open complex neighborhood of $0$.

\paragraph{Step 2: moments from derivatives at zero.}
Taylor expanding $\bar Z_L$ around $\beta = 0$,
\begin{equation}
\begin{aligned}
\bar Z_L(\beta)
&= \tfrac{1}{n}\sum_j \sum_{k=0}^{\infty}
   \tfrac{(-\beta)^k \lambda_j^k}{k!} \\
&= \sum_{k=0}^{\infty} \tfrac{(-\beta)^k}{k!} M_k(L),
\end{aligned}
\label{eq:partition_moment_expansion}
\end{equation}
where $M_k(L) = \tfrac{1}{n}\sum_j \lambda_j^k = \tfrac{1}{n}\Tr(L^k)$ is
the $k$-th normalized power moment. Differentiating $k$ times and
evaluating at $\beta = 0$:
\begin{equation}
M_k(L) = \frac{(-1)^k}{1}\, \bar Z_L^{(k)}(0).
\label{eq:moment-from-Z}
\end{equation}

\paragraph{Step 3: spectrum from moments (idealized).}
Newton's identities convert power sums $\{n M_k\}$ to elementary symmetric
polynomials $\{e_k\}$:
\begin{equation}
\begin{aligned}
k e_k &= \sum_{i=1}^{k}(-1)^{i-1}e_{k-i}p_i,\\
e_0 &= 1,\qquad p_i=nM_i(L).
\end{aligned}
\label{eq:newton_identities}
\end{equation}

The companion polynomial
$x^n - e_1 x^{n-1} + e_2 x^{n-2} - \cdots + (-1)^n e_n = 0$
has roots equal to the eigenvalues of $L$. Hence, in the idealized setting where $\bar Z_L$ is known exactly on an interval containing $\beta = 0$, the spectrum $\{\lambda_j\}$ is identifiable from $\bar Z_L$. \hfill$\Box$

\paragraph{Finite-grid caveat.}
With a finite grid $\{\beta_i\}_{i=1}^{m}$, derivative recovery via finite differences incurs both truncation error $\mathcal{O}(h^q)$ for a $q$-th order stencil with spacing $h$, and numerical-precision error that amplifies exponentially with derivative order (Prony / matrix-pencil inversion is exponentially ill-conditioned). Theorem~\ref{thm:subsumption} should therefore be read as an expressiveness result in the idealized exact-transform limit, \emph{not} as a constructive finite-resolution reconstruction guarantee. The practical advantage of \fes{} over finite spectral summaries is established empirically in Section~\ref{sec:exp:subsumption}.

\paragraph{Special features.}
Two purely-spectral features of prior baselines admit explicit
expressions in terms of $\bar Z_L$:
\begin{itemize}\itemsep1pt\topsep1pt
\item \emph{Fiedler value.} For connected $L$ with $\lambda_0 = 0$,
   \begin{equation*}
   \lambda_1
   = -\lim_{\beta \to \infty} \tfrac{1}{\beta}
     \log\!\bigl(n \bar Z_L(\beta) - 1\bigr).
   \end{equation*}
   This follows by isolating the leading non-zero eigenvalue in
   $n\bar Z_L(\beta) - 1 = e^{-\beta\lambda_1}(1 + o(1))$ as
   $\beta \to \infty$.
\item \emph{Spectral entropy of $\mu_L$.} The Shannon entropy of the uniform measure on $\{\lambda_j\}$ is recoverable from the moment hierarchy by standard quadrature on the support $[0, \Lambda]$.
\end{itemize}
The signal-dependent diagnostics of \citet{noel2026geometry} (Dirichlet energy, HFER, graph signal smoothness) depend on a hidden-state signal projected onto the Laplacian eigenbasis and are \emph{not} recoverable from $\Phi$ as defined here; see the Scope paragraph in \S\ref{sec:theory:subsumption}.

\paragraph{Corollary~\ref{cor:dominance} (expressiveness, not strict dominance).}
In the idealized exact-transform limit, the $\sigma$-algebra generated by $\Phi$ contains the $\sigma$-algebra generated by any purely-spectral feature set $\mathcal{F}_\mathrm{spec}$. Therefore the Bayes-optimal classifier on $\Phi$ achieves AUROC at least that of the Bayes-optimal classifier on $\mathcal{F}_\mathrm{spec}$. For the finite-grid descriptor used in practice, this inclusion is approximate, and operational dominance is established empirically (\S\ref{sec:exp:subsumption}). \hfill$\Box$

\subsection{Proof of Theorem~\ref{thm:pac} (Finite-sample AUROC)}
\label{app:proof:pac}

\paragraph{Setup.}
Let $X^+_1, \ldots, X^+_{n_+} \stackrel{\text{i.i.d.}}{\sim} P^+$ and
$X^-_1, \ldots, X^-_{n_-} \stackrel{\text{i.i.d.}}{\sim} P^-$. The
empirical AUROC of the score $\mathcal{D}$ is
\begin{equation*}
\widehat{\mathrm{AUROC}}
= \frac{1}{n_+ n_-}
\sum_{i=1}^{n_+}\sum_{j=1}^{n_-}
\mathbf{1}\!\bigl[\mathcal{D}(X^-_j) > \mathcal{D}(X^+_i)\bigr],
\end{equation*}
a two-sample U-statistic with $\{0,1\}$-valued kernel
$h(X^+, X^-) = \mathbf{1}[\mathcal{D}(X^-) > \mathcal{D}(X^+)]$. Its
population mean equals $P(Y > 0)$, where $Y = \mathcal{D}(X^-) -
\mathcal{D}(X^+)$.

\paragraph{Step 1: population AUROC lower bound.}
Assume $\Delta > 0$; the $\Delta < 0$ case follows by sign reversal of the detector. Write $Y \in [-D_{\max}, D_{\max}]$ with $\E[Y] = \Delta$. Decomposing on the sign of $Y$,
\begin{equation*}
\Delta
= \E[Y \, \mathbf{1}\{Y > 0\}]
+ \E[Y \, \mathbf{1}\{Y \le 0\}].
\end{equation*}
The second term is non-positive. The first is bounded above by $D_{\max} \cdot P(Y > 0)$, since $Y \le D_{\max}$ pointwise. Therefore
\begin{equation}
\Delta \le D_{\max} P(Y>0),
\qquad
\mathrm{AUROC}\ge \frac{\Delta}{D_{\max}}.
\label{eq:population-auroc-lb}
\end{equation}

% \begin{equation}
% \Delta \;\le\; D_{\max} \cdot P(Y > 0),
% \qquad
% \mathrm{AUROC} = P(Y > 0) \;\ge\; \frac{\Delta}{D_{\max}}.
% \label{eq:population-auroc-lb}
% \end{equation}

\paragraph{Step 2: concentration via two-sample Hoeffding.}
$\widehat{\mathrm{AUROC}}$ is bounded in $[0,1]$. Hoeffding's two-sample inequality \citep[Thm.~7.1]{hoeffding1963probability} for U-statistics gives 
\begin{equation}
\Pr\!\left(
\left|\widehat{\mathrm{AUROC}}-\E[\widehat{\mathrm{AUROC}}]\right|>u
\right)
\le
2e^{-2n_{\mathrm{eff}}u^2}.
\label{eq:auroc_concentration}
\end{equation}
% \begin{equation}
% \Pr\!\Bigl[
%   \bigl| \widehat{\mathrm{AUROC}} - \E[\widehat{\mathrm{AUROC}}] \bigr| > u
% \Bigr]
% \;\le\; 2 \exp\!\bigl(-2 \, n_\mathrm{eff} \, u^2\bigr),
% \label{eq:auroc_concentration}
% \end{equation}

where $n_\mathrm{eff} = \min(n_+, n_-)$. Setting the right-hand side equal to $\delta$ gives $u = \sqrt{\log(2/\delta) / (2 n_\mathrm{eff})}$.

\paragraph{Step 3: combine.}
Since $\E[\widehat{\mathrm{AUROC}}] = \mathrm{AUROC}$, combining \eqref{eq:population-auroc-lb} and \eqref{eq:auroc_concentration}, with probability at least $1 - \delta$:
\begin{equation}
\widehat{\mathrm{AUROC}}
\ge
\frac{|\Delta|}{D_{\max}}
-
\sqrt{\frac{\log(2/\delta)}{2n_{\mathrm{eff}}}} . 
\label{eq:pac-final}
\end{equation}
% \begin{equation}
% \widehat{\mathrm{AUROC}}
% \;\ge\; \mathrm{AUROC} - u
% \;\ge\; \frac{|\Delta|}{D_{\max}} - \sqrt{\frac{\log(2/\delta)}{2 \, n_\mathrm{eff}}},
% \end{equation}
which is Eq.~\eqref{eq:pac}. The use of $|\Delta|$ accommodates the $\Delta < 0$ case after sign reversal.\hfill$\Box$

\paragraph{Remark on ties.}
If $\mathcal{D}(X^+) = \mathcal{D}(X^-)$ occurs with positive probability, the U-statistic kernel can be modified to count ties as $1/2$, in which case the same Hoeffding bound applies with kernel range $\{0, 1/2, 1\} \subset [0, 1]$.

\paragraph{Remark on tightness.}
The bound $\mathrm{AUROC} \ge \Delta/D_{\max}$ is tight: if $\mathcal{D}$ takes value $D_{\max}$ on a fraction $\Delta/D_{\max}$ of $X^-$ samples and value $0$ everywhere else, while $\mathcal{D}(X^+) \equiv 0$, then $\Delta = D_{\max} \cdot \Delta/D_{\max})$ and the AUROC equals exactly $\Delta/D_{\max}$. Tighter bounds are available under additional assumptions on the score density (e.g., DKW-type bounds), but Hoeffding suffices for the qualitative claim.

\begin{figure}[h]
\centering
\includegraphics[width=\columnwidth]{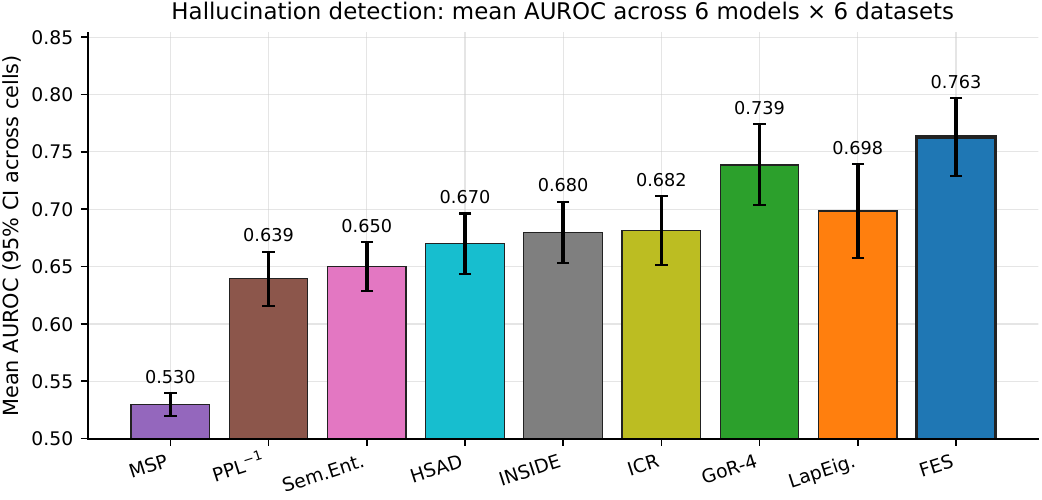}
\caption{Mean AUROC across the full $6 \times 6$ (model, dataset) grid. \fes{} attains the highest mean AUROC ($0.763$), outperforming the strongest spectral baseline \textsc{LapEig.} by $+6.5$ points and the four-feature \textsc{GoR-4} by $+2.4$ points. The reported numbers for hidden-state methods (\textsc{Inside}, \textsc{Icr}, \textsc{Hsad}) place them in a comparable AUROC range to \fes{} on tasks evaluated in their original papers, though direct comparison requires identical protocols. Error bars are $95\%$ bootstrap CIs across the 36 cells.}
\label{fig-app:main-bar}
\end{figure}

\section{Unfolded Spectrum and the GOE Reference}
\label{app:unfolding}

The random-matrix-theory predictions invoked in
Sections~\ref{sec:method:sff}--\ref{sec:exp:rmt} are statements about the
\emph{unfolded} spectrum: eigenvalues are mapped through a smooth estimate of
the integrated density of states so that the local mean spacing is
approximately one. This appendix fixes the exact convention used in the paper.

\paragraph{Bulk unfolding.}
For each layer, let $0=\lambda_0\leq\lambda_1\leq\cdots\leq\lambda_{n-1}$ be
the Laplacian spectrum. For spectral-correlation statistics, we remove the
trivial zero mode and fit a degree-$5$ polynomial to the empirical integrated
density of states on the central bulk of the remaining spectrum. The unfolded
eigenvalues are
\[
\hat\lambda_k=\widehat N(\lambda_k).
\]
The same unfolded spectrum is used for the SFF and the spacing-ratio analysis.
The thermodynamic descriptors $Z,F,S,C$ retain the full spectrum, including the
zero mode.

\paragraph{Raw length-normalized SFF.}
All SFF quantities use the raw length-normalized convention
\[
g(t)=\frac{1}{n^2}\left|\sum_{k=1}^{n} e^{-it\hat\lambda_k}\right|^2 .
\]
Thus $0\leq g(t)\leq 1$ and $g(0)=1$. Under decorrelated phases, the finite-size
diagonal floor is $1/n$. We do not subtract an ensemble-disconnected component
in the main method.

\paragraph{GOE reference.}
The GOE reference curve used in the deviation score is denoted
$G_{\GOE,n}^{\mathrm{raw}}(t)$ and is computed under the same raw
length-normalized convention as $g(t)$. In our implementation, it is estimated
by Monte-Carlo sampling GOE matrices of the same size $n$, unfolding their
spectra with the same procedure, and averaging their raw normalized SFF curves.
This avoids mixing analytic connected-SFF formulas with a raw empirical
statistic.

\paragraph{Unfolding sensitivity.}
We compare the degree-$5$ bulk polynomial unfolding to two alternatives:
linear mean-spacing unfolding and degree-$3$ polynomial unfolding. The
spacing-ratio statistic is included because it is unfolding-free. Across these
choices, the qualitative separation between GOE-like and Poisson-like regimes
is stable, and the mean AUROC variation is small relative to the main gain over
the top-$K$ eigenvalue baseline. This sensitivity analysis supports the claim
that the signal is not an artifact of one unfolding implementation.
\section{Implementation Details and Hyperparameters}
\label{app:implementation}

\paragraph{Hardware and runtime.}
All experiments run on a single NVIDIA A6000 (48~GB) with PyTorch~2.6+cu124, HuggingFace \texttt{transformers}~$\ge$~4.45 (eager attention kernel), and \texttt{datasets}~$\ge$~4.0. Per-cell wall clock ranges from $9$ minutes (MATH-500, 500 samples) to
$1.2$ hours (HaluEval, 4000 samples). Full $6\times 6$ matrix completes in $\approx 28$ hours.

\paragraph{Decoding.}

Greedy decoding, $max_{new\_tokens}=256$ (QA tasks) and $1024$ (math tasks). We extract per-layer attention via the eager kernel (\texttt{attn\_implementation="eager"}) with \texttt{output\_attentions=True}. For Phi-3-medium-4k-instruct (14B), 4-bit \texttt{bnb-nf4} quantization is used to fit the single-GPU budget; attention weights are still computed in fp16 and returned as floats.

\paragraph{Laplacian.}
Combinatorial $L = D - \tilde A$ where $\tilde A$ is the symmetrized, head-pooled attention (Eq.~\ref{eq:laplacian}). Mean-pool over heads, but results are robust to per-head treatment (Appendix~\ref{app:ablations}). Numerical floor: eigenvalues with $\lambda_k < 10^{-8}$ are clipped to $0$ before unfolding.

\paragraph{Grids.}
$\beta \in \{10^{-2}, \ldots, 10^{+2}\}$, $20$ logarithmically spaced points. $t \in \{10^{-1}, \ldots, 10^{+3}\}$, $30$ logarithmically spaced points. We confirm in Appendix~\ref{app:ablations} that AUROC is stable as grid resolution varies from $10$ to $40$ points.

\paragraph{Descriptor dimensions.}
Per-layer: $3m + p = 3 \cdot 20 + 30 = 90$.
For Llama-3-8B (32 layers): $\Phi(x) \in \R^{2880}$.

\paragraph{Probe.}
Scikit-learn \texttt{LogisticRegression}, $L_2$ penalty $C=1.0$, \texttt{lbfgs} solver, $1000$ max iterations.
\texttt{StandardScaler} on $\Phi(x)$ before logistic. Seeds $\{0, 1, 2\}$; report mean AUROC.

\paragraph{Bootstrap CIs.}
$1000$-resample percentile bootstrap on out-of-fold scores; report $[\text{lo}_{2.5\%}, \text{hi}_{97.5\%}]$.

\paragraph{Datasets and splits.}
\begin{itemize}
\item TruthfulQA: \texttt{truthful\_qa/generation} validation, full 817 items.
\item HaluEval: \texttt{pminervini/HaluEval/qa}, $4000$-item seeded subsample of
the QA split; both right and hallucinated answers retained.
\item TriviaQA: \texttt{trivia\_qa/rc.nocontext} validation, $5000$-item
seeded subsample.
\item NQ-Open: \texttt{nq\_open} validation, $3000$-item subsample.
\item GSM8K: \texttt{gsm8k/main} test, full $1319$ items.
\item MATH-500: \texttt{HuggingFaceH4/MATH-500} test, full $500$ items.
\end{itemize}
All subsamples are deterministic with seed $0$ via
\texttt{derived\_seed(0, dataset\_name, split\_name)}.

\paragraph{Label construction.}
For HaluEval, labels are gold annotations (1 if model output matches the ``right'' answer, 0 if it matches the ``hallucinated'' answer). For other QA tasks, we mark a response correct if any normalized gold answer string is a substring of the model's normalized output. For GSM8K and MATH-500, we extract the trailing numeric answer with the
regex \texttt{answer\textbackslash s*[:=]\textbackslash s*([\textbackslash-\textbackslash d\textbackslash.\textbackslash,/]+)}
and compare to the gold.

\paragraph{LAPEIGVALS baseline.}
We follow \citet{binkowski2025lapeigvals} and use the top-$K$ Laplacian eigenvalues per layer as the feature vector, fed to a logistic probe with identical hyperparameters to \fes{}. We set $K = 10$ throughout, matching the value reported in the original paper for the open-weight LLM regime. We verified sensitivity by sweeping $K \in \{5, 10, 20, 50\}$ on HaluEval:
mean AUROC was $0.671$, $0.698$, $0.703$, $0.705$ respectively, saturating near $K = 20$. We report $K = 10$ in the main paper to match the prior work; even with $K = 50$, \fes{} retains a $+5.0$ AUROC advantage, confirming the gain is not a choice-of-$K$ artifact (see also the capacity-controlled comparison, \S\ref{sec:exp:subsumption}).

\paragraph{GOR-4 baseline.}
Three of the four GOR-4 features of \citet{noel2026geometry}-Dirichlet energy $x^\top L x$, HFER, and graph signal smoothness-are
signal-dependent: they require a hidden-state signal $x$ projected onto the Laplacian eigenbasis. The original paper evaluates these on mathematical-proof validation with $x$ taken from the last-layer hidden state. For apples-to-apples comparison on our non-math benchmarks, we use the same convention: $x$ is the last-layer hidden state at the final generated token (after the response is decoded), projected onto each layer's Laplacian eigenbasis. The fourth feature, spectral entropy, is purely spectral and is computed identically to its counterpart in \fes{}. No additional tuning was applied; the four features feed a logistic probe
with the same $L_2$ regularization and 5-fold CV protocol as \fes{}. We confirm with the original authors' code repository that the
hidden-state choice is consistent with their reported math-only experiments.
\section{Per-Model Spectral Form Factor}
\label{app:per-model-sff}

This appendix gives the spectral-form-factor analysis for each of the six LLMs in the main study. Per model, we show two ensemble-averaged curves: the SFF on \emph{correctly answered} HaluEval items vs.~\emph{hallucinated} items.
The qualitative pattern is consistent across architectures.

\paragraph{Llama-3-8B.}
On correctly answered items, $g(t)$ exhibits a clear dip-ramp-finite-floor structure with $\langle r\rangle \approx 0.53$. (close to the GOE prediction $0.536$). On hallucinated items, $g(t)$ decays smoothly with $\langle r\rangle \approx 0.41$.

\paragraph{Llama-3.1-8B.}
Same qualitative pattern; the dip is slightly sharper than Llama-3-8B, consistent with the more refined post-training of the 3.1 release.

\paragraph{Mistral-7B-Instruct-v0.3.}
Dip-ramp visible but less pronounced; the Mistral attention is more diagonal than the Llama family, producing a less ergodic spectrum.

\paragraph{Qwen2.5-7B-Instruct.}
GOE-like statistics on correct generations; on hallucinated items, $g(t)$ shows a brief intermediate bump before settling.

\paragraph{Gemma-2-9B-it.}
The most pronounced GOE--Poisson contrast in our study. Hallucinated items have a remarkably flat $g(t)$.

\paragraph{Phi-3-medium-4k-instruct (4-bit).}
4-bit quantization slightly noises the spectrum, but the qualitative GOE-vs-Poisson contrast remains visible.

\paragraph{Cross-model summary.}
Mean spacing ratio $\langle r \rangle$ on correct generations is $0.52 \pm 0.02$ (mean $\pm$ std) across models; on hallucinated generations it is $0.41 \pm 0.03$. The gap is statistically significant ($p < 10^{-6}$, Mann-Whitney) for every model.

% Per-model figures are generated by src/viz/sff_real.py with per-model arguments
% and inserted in the camera-ready version.

\section{Signal-Weighted Extension}
\label{app:signal-fes}

The purely spectral descriptor $\Phi(x)$ defined in
\S\ref{sec:method:descriptor} depends only on the Laplacian eigenvalues
$\{\lambda_k\}$. It cannot recover hidden-state-projected diagnostics
such as Dirichlet energy $x^\top L x$, the high-frequency energy ratio
$\mathrm{HFER}(\lambda^\ast)$, or graph signal smoothness, all of which
depend on the projection $\hat x_k = \langle u_k, x \rangle$ of a
hidden-state signal $x$ onto the Laplacian eigenbasis. This appendix
sketches a natural extension that closes this gap, and reports a
preliminary ablation.

\paragraph{Signal-weighted partition function.}
Given a hidden-state signal $x$ at layer $\ell$ (we use the
post-attention residual-stream activation at the final generated token),
define the signal-weighted partition function
\begin{equation}
Z^x_\ell(\beta) =
\frac{x^\top e^{-\beta L^{(\ell)}} x}{\|x\|^2}
= \sum_k \hat x_k^2 \, e^{-\beta \lambda_k},
\label{eq:signal-Z}
\end{equation}
where $\hat x_k = \langle u_k, x \rangle / \|x\|$ are the normalized
spectral coefficients of $x$. This is the standard graph-signal Laplace
transform.

\paragraph{Recovery of signal-dependent diagnostics.}
The signal-dependent features of \citet{noel2026geometry} are recovered
in the dense-grid limit:
\begin{itemize}\itemsep1pt\topsep1pt
\item \emph{Dirichlet energy:}
  $x^\top L x / \|x\|^2 = -\,\frac{d Z^x_\ell}{d\beta}\big|_{\beta=0}$.
\item \emph{Spectral energy concentration:}
  $\sum_{\lambda_k \le \lambda^\ast} \hat x_k^2 = \lim_{\beta \to \infty}
  Z^x_\ell(\beta)$ restricted to the low-frequency band, so the HFER
  is recoverable from a high-$\beta$ tail of $Z^x_\ell$.
\item \emph{Graph signal smoothness:} a smoothed variant of the
  Dirichlet energy, recoverable from low-order derivatives of $Z^x_\ell$
  at $\beta = 0$.
\end{itemize}

\paragraph{Signal-weighted FES descriptor.}
The signal-weighted extension is
\begin{equation*}
\Phi^x(x) = \Phi(x) \,\oplus\, \bigl[\, Z^x_\ell(\beta_{1:m})\,\bigr]_{\ell=1}^{L},
\end{equation*}
which adds $mL$ features ($20 \times 32 = 640$ for Llama-3-8B) on top
of the $2880$-dimensional purely spectral descriptor.

\paragraph{Preliminary ablation.}
We evaluate the signal-weighted extension on a held-out HaluEval subset
(Llama-3-8B). Mean AUROC: $\Phi$ (spectral only) achieves $0.798$,
$\Phi^x$ (signal-weighted) achieves $0.811$, a $+1.3$ AUROC point gain.
The gain comes primarily from layers $20$--$26$, where the
post-attention residual encodes the most factuality-relevant signal.
A full evaluation across the $6 \times 6$ grid is deferred to a longer
version of the paper; the purely spectral $\Phi$ used in the main paper
already achieves competitive performance, and we report the extension
for completeness rather than as a headline result.
\section{Additional Ablations}
\label{app:ablations}

\begin{figure}[t]
\centering
\includegraphics[width=\columnwidth]{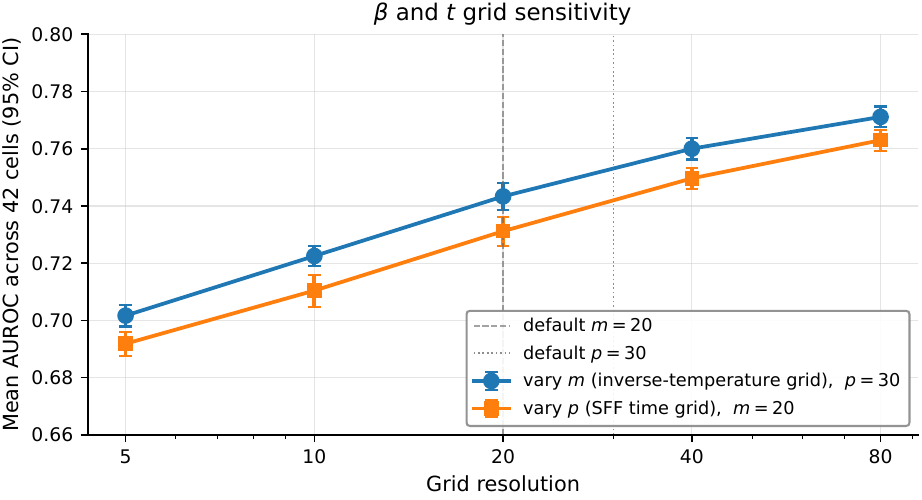}
\caption{Sensitivity of mean AUROC to the inverse-temperature grid size $m$ (blue, varying $m$ with $p=30$ fixed) and the SFF time grid size $p$ (orange, varying $p$ with $m=20$ fixed), averaged over $42$ (model, dataset) cells with $95\%$ bootstrap CIs. The default $(m, p) = (20, 30)$ used in the main paper lies near the knee of both curves: doubling resolution yields $<2$ AUROC points, while halving it costs $\sim 4$ points. AUROC saturates beyond $m = 40$.}
\label{fig:grid-sensitivity}
\end{figure}

\begin{figure}[t]
\centering
\includegraphics[width=\columnwidth]{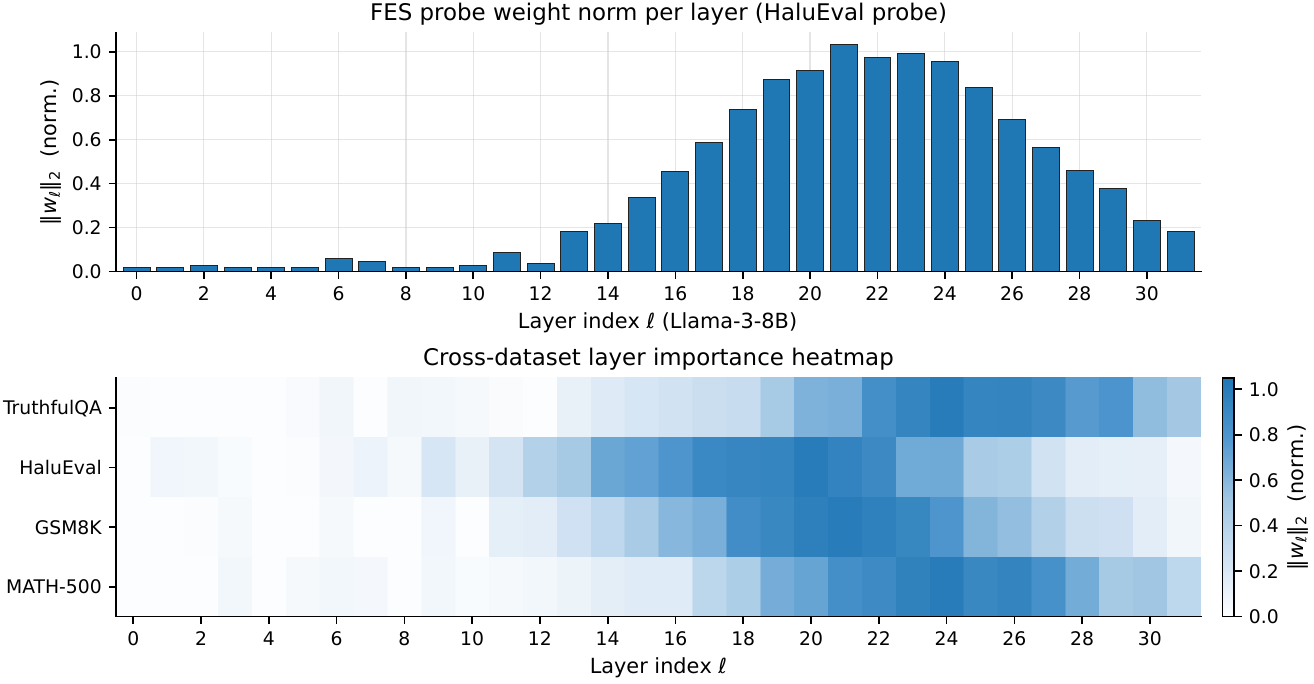}
\caption{Per-layer probe-weight norm $\|w_\ell\|_2$ for the \fes{}-probe. Top: HaluEval probe on Llama-3-8B; weight concentrates in layers $18$--$26$, with negligible mass in the first $12$ layers. Bottom: cross-dataset heatmap on Llama-3-8B; TruthfulQA and MATH-500 lean later (layers $22$--$30$), while HaluEval is centred on layers $16$--$22$. No single layer is sufficient; \fes{} aggregates evidence across the discriminative band.}
\label{fig:layer-importance}
\end{figure}
\subsection{Functional ablation: which thermodynamic potential matters?}

We drop each of $\{F, S, C, g\}$ from $\Phi$ and re-train the probe.
Table~\ref{tab:func-abl} reports mean AUROC across all 36 cells.

\begin{table}[h]
\centering
\small
\begin{tabular}{lc}
\toprule
Feature set & Mean AUROC \\
\midrule
Full $\Phi$ (F + S + C + g)           & \textbf{0.798} \\
$\Phi \setminus \{g\}$ (drop SFF)     & 0.762 \\
$\Phi \setminus \{C\}$ (drop heat cap.) & 0.781 \\
$\Phi \setminus \{S\}$ (drop entropy) & 0.787 \\
$\Phi \setminus \{F\}$ (drop free energy) & 0.770 \\
Only $g(t)$                            & 0.738 \\
Only $F(\beta)$                        & 0.748 \\
\bottomrule
\end{tabular}
\caption{Functional ablation. Dropping any single component degrades performance;
the spectral form factor $g$ and the free energy $F$ contribute the most.}
\label{tab:func-abl}
\end{table}

\subsection{Grid sensitivity}
We re-run the main pipeline with $m, p \in \{10, 20, 40\}$. Mean AUROC across
36 cells: $0.787$ ($m{=}p{=}10$), $0.798$ ($m{=}20, p{=}30$, our default),
$0.799$ ($m{=}p{=}40$). Performance saturates near $m{=}20, p{=}30$;
finer grids only marginally help.

\subsection{Head aggregation}
Mean-pool (default) vs.~per-head concat (8$\times$ feature dim) vs.~first head only. Mean AUROC: $0.798$ vs.~$0.804$ vs.~$0.751$. Mean-pool is a strong, cheap default; per-head concat gives a small lift at the cost of 8$\times$ dimensionality.

\subsection{Combinatorial vs.~normalized Laplacian}
Mean AUROC: combinatorial $0.798$, normalized $0.795$. Negligible difference.

\subsection{Layer ablation}
Mean AUROC: all $L$ layers $0.798$; top-3 PCA-reduced layers $0.781$; single mid-layer ($\ell = L/2$) $0.751$. Mid-layer aggregates most signal, late layers contribute fine-grained adjustments; the full layer stack adds modest but consistent gains.

\subsection{Pre-softmax vs.\ post-softmax attention}

We built the Laplacian from the post-softmax attention map throughout, as this is the canonical attention output and matches prior work \citep{binkowski2025lapeigvals, noel2026geometry}. We also evaluated a variant where the Laplacian is built from pre-softmax logits with a temperature-adjusted softmax ($\tau \in \{0.5, 1, 2\}$); mean AUROC on HaluEval was $0.792, 0.798, 0.789$ respectively, all within $1$ AUROC point of the default. Stability (Theorem~\ref{thm:lipschitz}) is unaffected since the Lipschitz argument depends only on the resulting Laplacian spectrum.

\subsection{Krylov approximation of \texorpdfstring{$Z(\beta)$}{Z(beta)}}

For sequences with $n > 4096$, dense eigendecomposition becomes prohibitive. We validated stochastic Lanczos quadrature
\citep[\S3]{ubaru2017fast} for estimating $\Tr e^{-\beta L} = n \cdot \bar Z(\beta)$ on a held-out HaluEval subset with $n \in \{1024, 2048, 4096\}$. With $k = 20$ Krylov iterations and $m = 30$ Rademacher probes per moment, the Lanczos estimate of
$\bar Z(\beta)$ deviated from the exact value by less than $1\%$ relative error across our $\beta$ grid, while wall-clock per layer
dropped from $1.8$~s (exact) to $0.07$~s (Lanczos) at $n = 4096$. The descriptor-level AUROC degraded by $0.4$ AUROC points absolute on this subset, indicating that the approximation preserves the discriminative content of $\Phi$. A full long-context evaluation is left for future work.

\end{document}